\documentclass[10pt,twocolumn,letterpaper]{article}

\usepackage[pagenumbers]{wacv} 

\usepackage{graphicx}
\usepackage{amsmath}
\usepackage{amssymb}
\usepackage{booktabs}

\usepackage{times}
\usepackage{epsfig}
\usepackage{subcaption}
\usepackage{multirow}
\usepackage{color,colortbl}

\definecolor{babyblueeyes}{rgb}{0.63, 0.79, 0.95}

\newcommand{\diag}{\mathop{\mathrm{diag}}}
\makeatletter
\def\blfootnote{\gdef\@thefnmark{}\@footnotetext}
\makeatother

\usepackage[pagebackref,breaklinks,colorlinks]{hyperref}

\usepackage[capitalize]{cleveref}
\crefname{section}{Sec.}{Secs.}
\Crefname{section}{Section}{Sections}
\Crefname{table}{Table}{Tables}
\crefname{table}{Tab.}{Tabs.}

\def\wacvPaperID{00048} 
\def\confName{WACV}
\def\confYear{2024}

\begin{document}

\title{Permutation-Aware Activity Segmentation via Unsupervised Frame-to-Segment Alignment}

\author{Quoc-Huy Tran$^*$~~~~~~~~~~Ahmed Mehmood$^*$~~~~~~~~~~Muhammad Ahmed~~~~~~~~~~Muhammad Naufil\\Anas Zafar~~~~~~~~~~Andrey Konin~~~~~~~~~~M. Zeeshan Zia\\
\\
Retrocausal, Inc., Redmond, WA\\
\url{www.retrocausal.ai}
}

\maketitle

\begin{abstract}
This paper presents an unsupervised transformer-based framework for temporal activity segmentation which leverages not only frame-level cues but also segment-level cues. This is in contrast with previous methods which often rely on frame-level information only. Our approach begins with a frame-level prediction module which estimates framewise action classes via a transformer encoder. The frame-level prediction module is trained in an unsupervised manner via temporal optimal transport. To exploit segment-level information, we utilize a segment-level prediction module and a frame-to-segment alignment module. The former includes a transformer decoder for estimating video transcripts, while the latter matches frame-level features with segment-level features, yielding permutation-aware segmentation results. Moreover, inspired by temporal optimal transport, we introduce simple-yet-effective pseudo labels for unsupervised training of the above modules. Our experiments on four public datasets, i.e., 50 Salads, YouTube Instructions, Breakfast, and Desktop Assembly show that our approach achieves comparable or better performance than previous methods in unsupervised activity segmentation. Our code and dataset are available on our research website: \url{https://retrocausal.ai/research/}.
\end{abstract}

\section{Introduction}
\label{sec:introduction}
{\blfootnote{$^*$ indicates joint first author.\\ \{huy,ahmedraza,ahmed,naufil,anas,andrey,zeeshan\}@retrocausal.ai.}}

\begin{figure}[t]
    \centering
	\includegraphics[width=0.92\linewidth, trim = 0mm 65mm 115mm 0mm, clip]{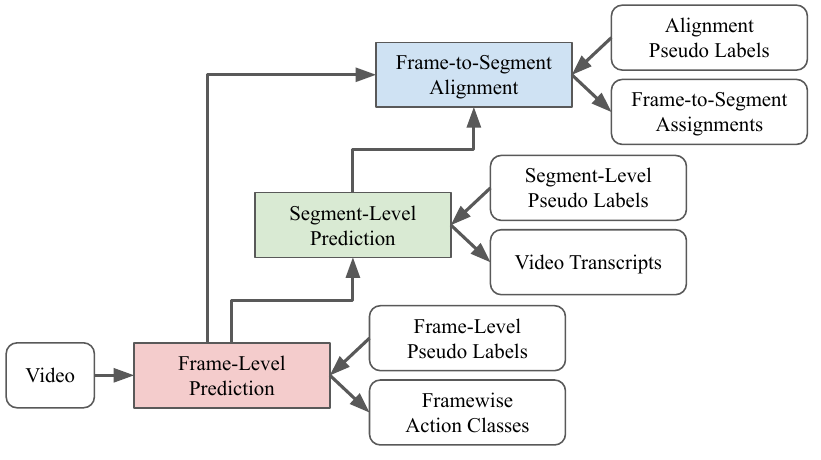}
    \caption{Prior works often use only frame-level cues via frame-level prediction modules (i.e., red) to predict framewise action classes. We adopt a segment-level prediction module and a frame-to-segment alignment module (i.e., green/blue), which exploit segment-level cues for permutation-aware results. Also, we introduce simple-yet-effective pseudo labels for unsupervised training.}
    \label{fig:teaser}
\end{figure}

Temporal activity segmentation~\cite{farha2019ms,li2020ms,chang2019d3tw,li2019weakly,rahaman2022generalized,behrmann2022unified,swetha2021unsupervised,kumar2022unsupervised,ding2022temporal} aims to associate each frame in a video capturing a human activity with one of the action/sub-activity classes. Temporally segmenting human activities in videos plays an important role in several computer vision, robotics, healthcare, manufacturing, and surveillance applications. Examples include visual analytics~\cite{akkas2017measuring,ji2020motion,ji2022computer} (i.e., compute time and motion statistics such as average cycle time from video recordings), ergonomics risk assessment~\cite{parsa2020spatio,parsa2021multi} (i.e., segment actions of interest in videos for analyzing ergonomics risks), and task guidance~\cite{funk2015benchmark,blattgerste2017comparing,ng2019using} (i.e., offer instructions to workers based on expert demonstration videos).

Considerable efforts have been made in designing fully-supervised methods~\cite{lea2017temporal,ding2017tricornet,lei2018temporal,farha2019ms,li2020ms} or weakly-supervised methods~\cite{kuehne2017weakly,richard2017weakly,richard2018neuralnetwork,ding2018weakly,chang2019d3tw,li2019weakly,li2021temporal,souri2022robust,khan2022timestamp,rahaman2022generalized,behrmann2022unified} for temporal activity segmentation due to their great performance. However, acquiring dense framewise labels or weak annotations such as transcripts~\cite{kuehne2017weakly} and timestamps~\cite{li2021temporal} is generally hard and expensive especially for a large number of videos. Therefore, we are interested in unsupervised approaches for temporal activity segmentation, which simultaneously extract actions and segment all video frames into clusters with each cluster representing one action. Early unsupervised methods~\cite{sener2018unsupervised,kukleva2019unsupervised,vidalmata2021joint,li2021action,wang2022sscap} separate representation learning from clustering, preventing effective feedback between them, while using offline clustering, resulting in memory inefficiency. To address that, UDE~\cite{swetha2021unsupervised} and TOT~\cite{kumar2022unsupervised} develop joint representation learning and online clustering approaches. The above methods often leverage frame-level information only (i.e., red block in Fig.~\ref{fig:teaser}), while not explicitly utilizing high-level information such as transcript, which is crucial for handling permutations of actions, missing actions, and repetitive actions.

In this work, we present an unsupervised activity segmentation framework which is based on transformers~\cite{vaswani2017attention} and exploits both frame-level cues and segment-level cues. Motivated by the strong performance of supervised transformer-based architectures~\cite{dosovitskiyimage,behrmann2022unified} in supervised activity segmentation, our unsupervised model includes a transformer encoder and a transformer decoder. The former performs self-attention to learn dependencies within the video sequence, while the latter relies on cross-attention to learn dependencies between the video sequence and the transcript sequence, resulting in effective contextual features. In addition to the frame-level prediction module for exploiting frame-level cues, we include a segment-level prediction module and a frame-to-segment alignment module (i.e., green and blue blocks in Fig.~\ref{fig:teaser}) to leverage segment-level cues, yielding permutation-aware segmentation results. For unsupervised training of the above modules, we propose simple-yet-effective pseudo labels based on temporal optimal transport~\cite{kumar2022unsupervised}. We demonstrate comparable or superior performance of our approach over previous unsupervised activity segmentation methods on four public datasets.

In summary, our contributions include:
\begin{itemize}
    \item We introduce a novel combination of modules and unsupervised losses to exploit both frame-level cues and segment-level cues for permutation-aware activity segmentation.
    \item We propose simple-yet-effective pseudo labels based on temporal optimal transport, enabling unsupervised training of the segment-level prediction module and the frame-to-segment alignment module.
    \item Extensive evaluations on 50 Salads, YouTube Instructions, Breakfast, and Desktop Assembly datasets show that our approach performs on par with or better than prior methods in unsupervised activity segmentation.
\end{itemize}

\section{Related Work}
\label{sec:relatedwork}

\noindent \textbf{Fully-Supervised Activity Segmentation.}
Early works in fully-supervised activity segmentation often rely on sliding temporal window with non-maximum suppression~\cite{rohrbach2012database,karaman2014fast} or structured temporal modeling via hidden Markov models~\cite{tang2012learning,kuehne2016end}, while recent methods are mostly based on temporal convolutional networks (TCNs)~\cite{lea2017temporal,ding2017tricornet,lei2018temporal,farha2019ms,li2020ms}. Lea et al.~\cite{lea2017temporal} develop the first TCN-based solution, which includes an encoder-decoder architecture with temporal convolutions and deconvolutions to capture long-range temporal dependencies. TricorNet~\cite{ding2017tricornet} replaces the above decoder by a bi-directional LSTM, while TDRN~\cite{lei2018temporal} employs deformable temporal convolutions instead. Since these methods downsample videos to a temporal resolution, they fail to capture fine-grained details. Thus, multi-stage TCNs~\cite{farha2019ms,li2020ms} are introduced to maintain a high temporal resolution. However, due to performing framewise prediction, the above methods suffer from over-segmentation. To address that, refinement techniques, e.g., graph-based reasoning~\cite{huang2020improving} and boundary detection~\cite{ishikawa2021alleviating}, are proposed.

\noindent \textbf{Weakly-Supervised Activity Segmentation.}
Weakly-supervised activity segmentation methods utilize different forms of weak labels, including the ordered list of actions appearing in the video, i.e., transcript supervision~\cite{kuehne2017weakly,richard2017weakly,richard2018neuralnetwork,ding2018weakly,chang2019d3tw,li2019weakly}, or the set of actions occurring in the video, i.e., set supervision~\cite{richard2018action,fayyaz2020sct,li2020set}. Recently, timestamp supervision~\cite{li2021temporal,souri2022robust,khan2022timestamp,rahaman2022generalized}, which requires labeling a single frame per action segment, has attracted research interests, since it has similar annotation costs as transcript supervision but it yields better results thanks to the additional approximate segment location information in timestamp labels. More recently, Behrmann et al.~\cite{behrmann2022unified} introduce a unified fully-supervised and timestamp-supervised method, achieving competitive results. The above methods need either framewise labels for full supervision or weak labels for weak supervision, whereas our approach does not.

\noindent \textbf{Unsupervised Activity Segmentation.}
Early attempts~\cite{sener2015unsupervised,alayrac2016unsupervised} in unsupervised activity segmentation often utilize the narrations accompanied with the videos, however, these narrations are not always provided. That motivates the development of methods with only visual inputs~\cite{sener2018unsupervised,kukleva2019unsupervised,vidalmata2021joint,li2021action,swetha2021unsupervised,wang2022sscap,kumar2022unsupervised}. Mallow~\cite{sener2018unsupervised} learns an appearance model and a temporal model of the activity in an alternating manner. CTE~\cite{kukleva2019unsupervised} first learns a temporal embedding and then clusters the embedded features with K-Means. To improve CTE, VTE~\cite{vidalmata2021joint} adds a visual embedding, while ASAL~\cite{li2021action} adds an action-level embedding. SSCAP~\cite{wang2022sscap} first uses a video-based self-supervised model for feature extraction and then performs co-occurrence action parsing to capture the temporal structure of the activity. The aforementioned methods separate representation learning from offline clustering, preventing effective feedback between
them, whereas we follow recent approaches, i.e., UDE~\cite{swetha2021unsupervised} and TOT~\cite{kumar2022unsupervised}, to perform joint representation learning and online clustering. Furthermore, unlike UDE~\cite{swetha2021unsupervised} and TOT~\cite{kumar2022unsupervised}, which exploit frame-level cues only, we propose modules for exploiting segment-level cues and pseudo labels for unsupervised training, yielding improved results.

\noindent \textbf{Transformers in Activity Segmentation.}
After successes of transformers~\cite{vaswani2017attention} in natural language processing, there has been a wide adoption of transformers in computer vision~\cite{dosovitskiyimage,chen2021crossvit,arnab2021vivit,bertasius2021space}. Transformers focus on attention mechanism to extract contextual information over the entire sequence. Recently, a few methods~\cite{yi2021asformer,behrmann2022unified} have applied transformers for temporal activity segmentation. ASFormer~\cite{yi2021asformer} consists of encoder blocks, each of which includes a dilated temporal convolution and a self-attention layer, and decoder blocks, where cross-attention is used to gather information from encoder blocks. Due to making framewise prediction, ASFormer suffers from over-segmentation. To address that, UVAST~\cite{behrmann2022unified} uses a transformer decoder to predict the transcript and exploit segment-level cues. In this work, we adopt the transformer encoder of ASFormer~\cite{yi2021asformer} and the transformer decoder of UVAST~\cite{behrmann2022unified}. However, our overall architecture is different from them. Also, they require labels for supervised training, whereas we propose pseudo labels for unsupervised training.
\section{Our Approach}
\label{sec:method}

\begin{figure*}[t]
	\centering
		\includegraphics[width=0.85\linewidth, trim = 0mm 15mm 25mm 0mm, clip]{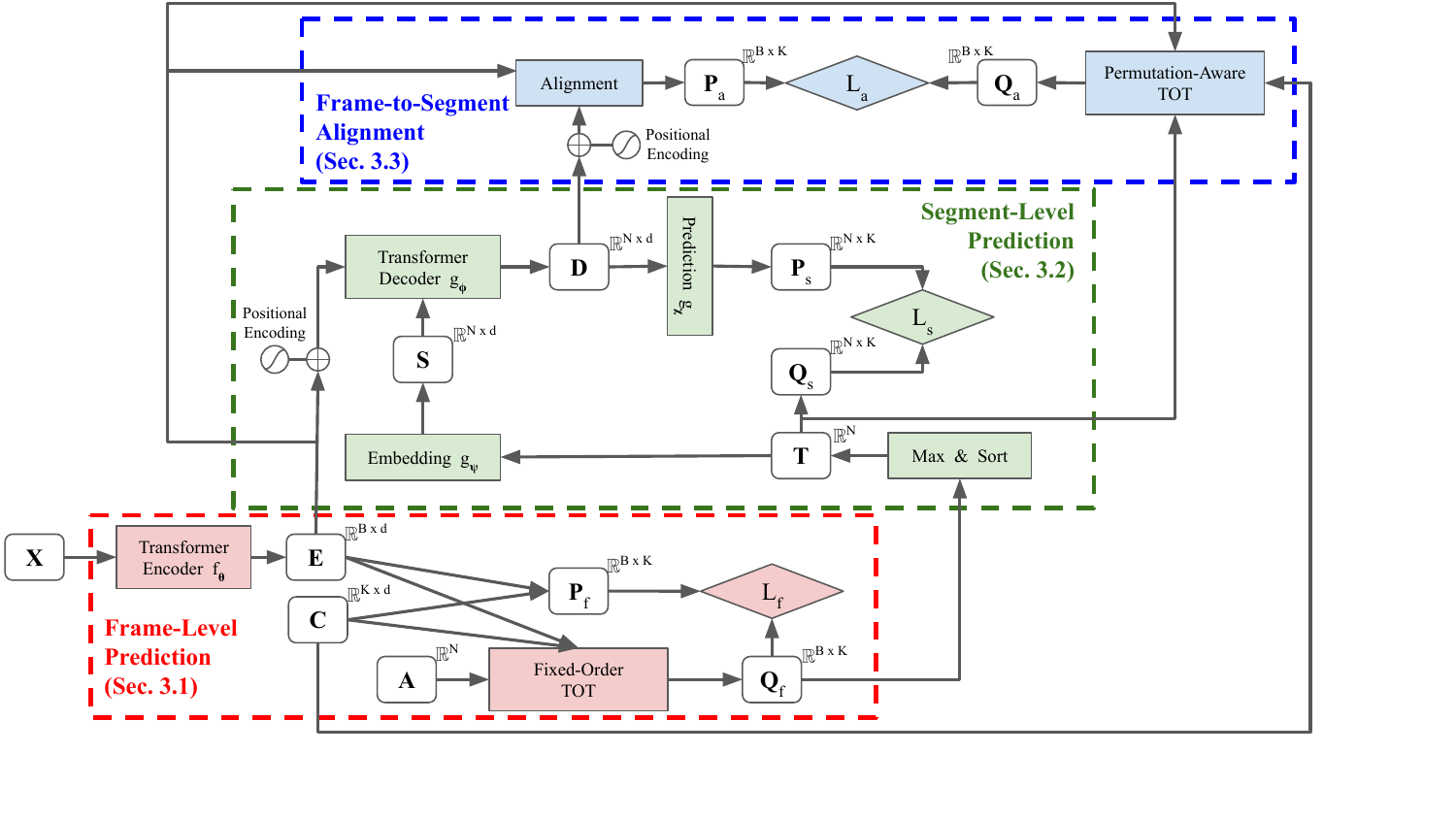}
	\caption{Our approach includes a frame-level prediction module (i.e., red) which extracts frame-level features $\boldsymbol{E}$ via a transformer encoder and uses temporal optimal transport to compute frame-level pseudo labels $\boldsymbol{Q}_f$ for unsupervised training. To exploit segment-level information, we utilize a segment-level prediction module (i.e., green), which extract segment-level features $\boldsymbol{D}$ via  a transformer decoder, and a frame-to-segment alignment module (i.e., blue), which matches frame-level features $\boldsymbol{E}$ and segment-level features $\boldsymbol{D}$. In addition, we introduce segment-level pseudo labels $\boldsymbol{Q}_s$ and alignment-level pseudo labels $\boldsymbol{Q}_a$ for unsupervised training of the above modules.}
	\label{fig:method}
\end{figure*}

We present below our main contribution, an unsupervised transformer-based framework for temporal activity segmentation. Fig.~\ref{fig:method} shows an overview of our approach.

\noindent \textbf{Notations.} Let us first represent the encoder function and the decoder function as $f_{\boldsymbol{\theta}}$ and $g_{\boldsymbol{\phi}}$ respectively (with learnable parameters $\boldsymbol{\theta}$ and $\boldsymbol{\phi}$). Our approach takes as input a sequence of $B$ frames, represented as $\boldsymbol{X} = [\boldsymbol{x}_1, \boldsymbol{x}_2, \dots, \boldsymbol{x}_B]^\top$. The encoder features of $\boldsymbol{X}$ are expressed as $\boldsymbol{E} = [\boldsymbol{e}_1, \boldsymbol{e}_2, \dots, \boldsymbol{e}_B]^\top \in \mathbb{R}^{B \times d}$ with $\boldsymbol{e}_i = f_{\boldsymbol{\theta}}(\boldsymbol{x}_i) \in \mathbb{R}^{d}$ ($d$ is the feature dimension). Next, let us denote $\boldsymbol{A} = [1,2,\dots,K]^\top \in \mathbb{R}^{K}$ as the sequence of $K$ action classes in the activity. Our approach learns a group of $K$ prototypes, represented as $\boldsymbol{C} = [\boldsymbol{c}_1, \boldsymbol{c}_2, \dots, \boldsymbol{c}_K]^\top \in \mathbb{R}^{K \times d}$ with $\boldsymbol{c}_j  \in \mathbb{R}^{d}$ corresponding to the $j$-th action class in $\boldsymbol{A}$. We denote $\boldsymbol{T} = [a_1,a_2,\dots,a_N]^\top \in \mathbb{R}^{N}$ (with $a_j \in \boldsymbol{A}$) as the transcript which contains the sequence of actions appearing in $\boldsymbol{X}$, and $\boldsymbol{S} \in \mathbb{R}^{N \times d}$ as the transcript features. The decoder features are written as $\boldsymbol{D} = [\boldsymbol{d}_1, \boldsymbol{d}_2, \dots, \boldsymbol{d}_N]^\top \in \mathbb{R}^{N \times d}$ with $\boldsymbol{d}_j  \in \mathbb{R}^{d}$ corresponding to $a_j$ in $\boldsymbol{T}$. Finally, we represent $\boldsymbol{P}_f \in \mathbb{R}^{B \times K}$, $\boldsymbol{P}_s \in \mathbb{R}^{N \times K}$, and $\boldsymbol{P}_a \in \mathbb{R}^{B \times K}$ as the \emph{predicted} assignment probabilities (i.e., predicted ``codes'') at the frame-level prediction module (i.e., between frames and actions), the segment-level prediction module (i.e., between transcript positions and actions), and the frame-to-segment alignment module (i.e., between frames and actions) respectively. Similarly, $\boldsymbol{Q}_f \in \mathbb{R}^{B \times K}$, $\boldsymbol{Q}_s \in \mathbb{R}^{N \times K}$, and $\boldsymbol{Q}_a \in \mathbb{R}^{B \times K}$ denote the corresponding \emph{pseudo-label} assignment probabilities (i.e., pseudo-label ``codes'') for $\boldsymbol{P}_f$, $\boldsymbol{P}_s$, and $\boldsymbol{P}_a$ respectively.

\subsection{Unsupervised Frame-Level Prediction}
\label{sec:framelevel}

\begin{figure}[t]
     \centering
     \begin{subfigure}[b]{0.21\textwidth}
         \centering
         \includegraphics[width=1.0\linewidth, trim = 0mm 15mm 100mm 0mm, clip]{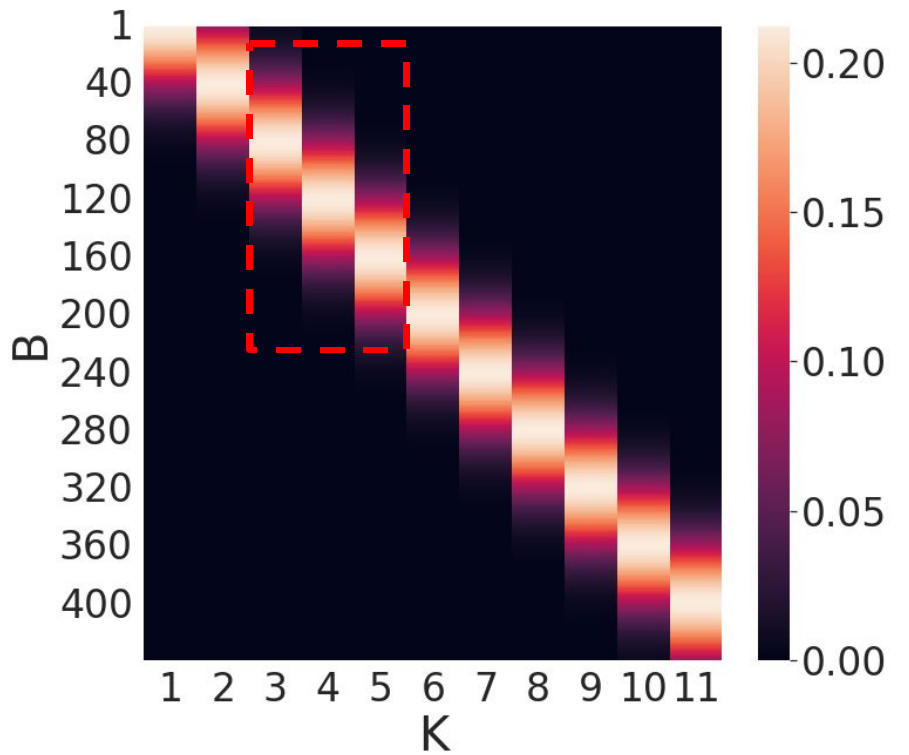}
         \caption{$\boldsymbol{M}_{\boldsymbol{A}}$ }
         \label{fig:M_A}
     \end{subfigure}
     \hfill
     \centering
     \begin{subfigure}[b]{0.21\textwidth}
         \centering
         \includegraphics[width=1.0\linewidth, trim = 0mm 14mm 100mm 0mm, clip]{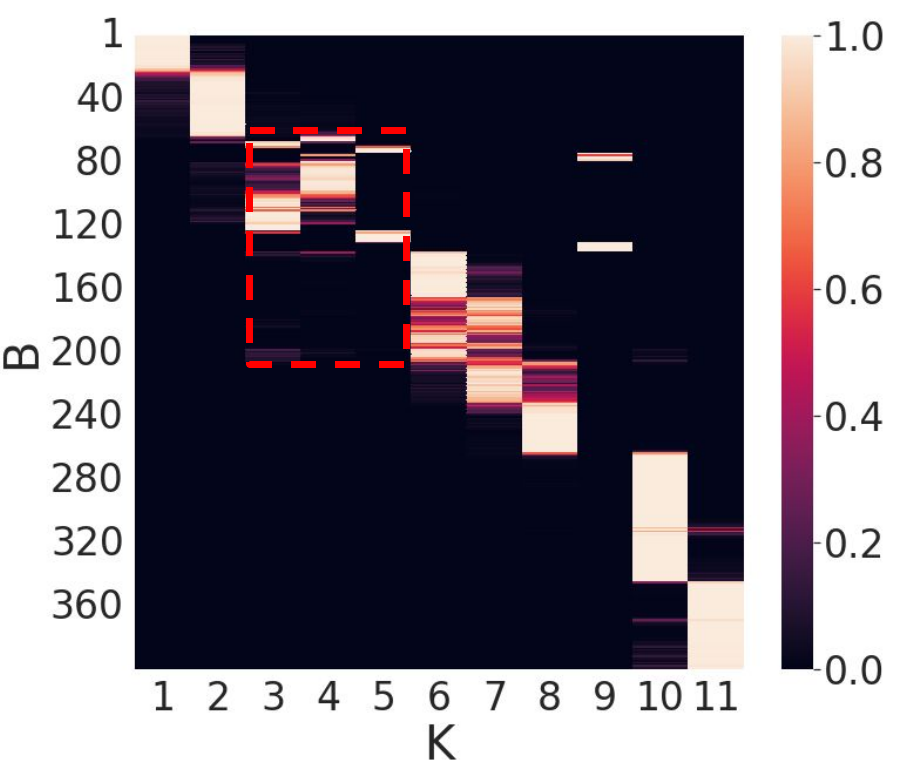}
         \caption{$\boldsymbol{Q}_f$}
         \label{fig:q_mat_f}
     \end{subfigure}
    \caption{(a) Fixed-order prior distribution $\boldsymbol{M}_{\boldsymbol{A}}$. (b) Frame-level pseudo-label codes $\boldsymbol{Q}_f$.}
    \label{fig:framelevel}
\end{figure}

\begin{figure}[t]
     \centering
     \begin{subfigure}[b]{0.21\textwidth}
         \centering
         \includegraphics[width=1.0\linewidth, trim = 0mm 15mm 100mm 0mm, clip]{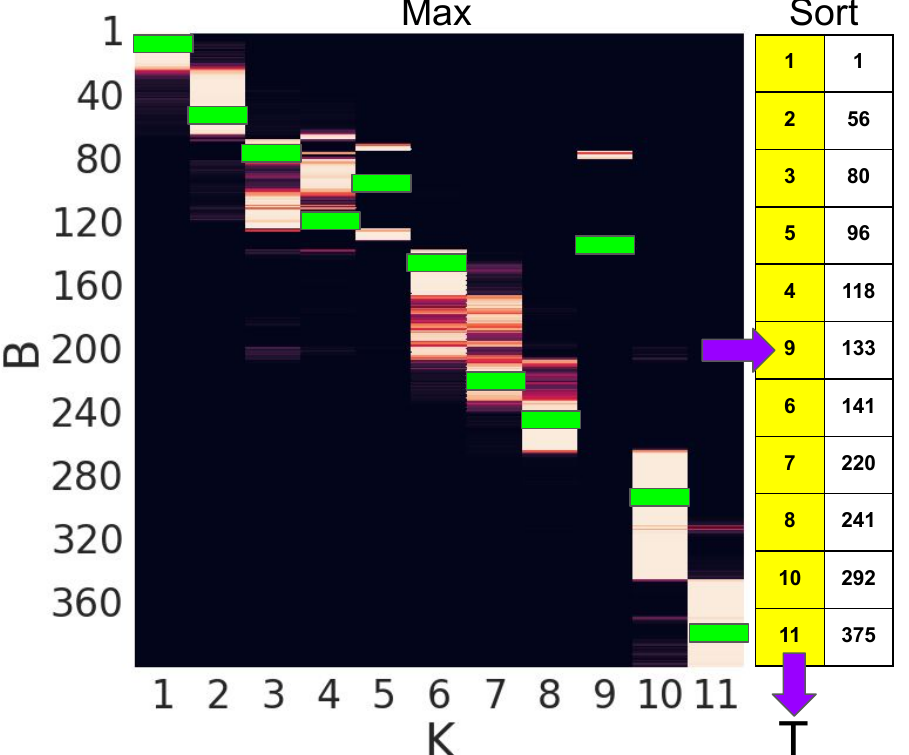}
         \caption{$\boldsymbol{T}$}
         \label{fig:transcript}
     \end{subfigure}
     \hfill
     \begin{subfigure}[b]{0.21\textwidth}
         \centering
         \includegraphics[width=1.0\linewidth, trim = 0mm 15mm 100mm 0mm, clip]{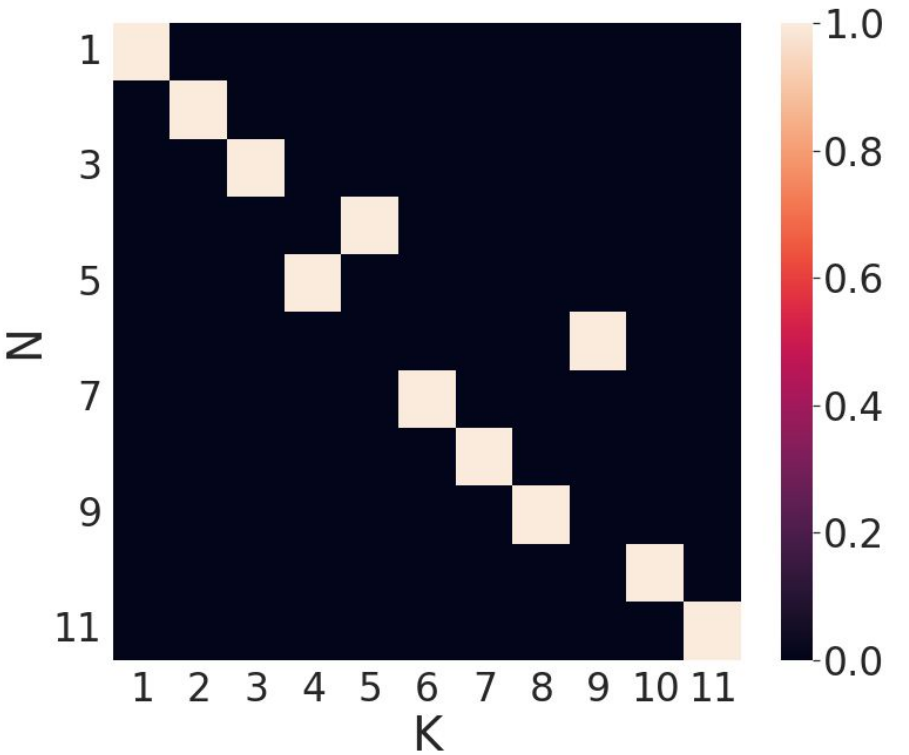}
         \caption{$\boldsymbol{Q}_s$}
         \label{fig:q_mat_s}
     \end{subfigure}
    \caption{(a) Permutation-aware transcript $\boldsymbol{T}$. (b) Segment-level pseudo-label codes $\boldsymbol{Q}_s$.}
    \label{fig:segmentlevel}
\end{figure}

\begin{figure}[t]
     \centering
     \begin{subfigure}[b]{0.21\textwidth}
         \centering
         \includegraphics[width=1.0\linewidth, trim = 0mm 15mm 100mm 0mm, clip]{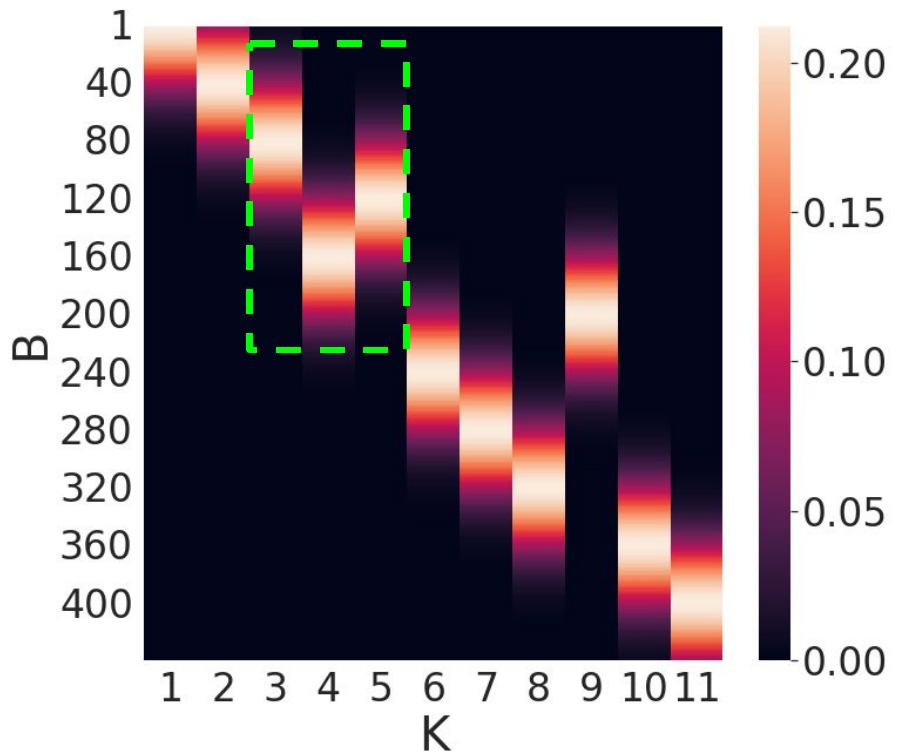}
         \caption{$\boldsymbol{M}_{\boldsymbol{T}}$ }
         \label{fig:M_T}
     \end{subfigure}
     \hfill
     \begin{subfigure}[b]{0.21\textwidth}
         \centering
         \includegraphics[width=1.0\linewidth, trim = 0mm 14mm 100mm 0mm, clip]{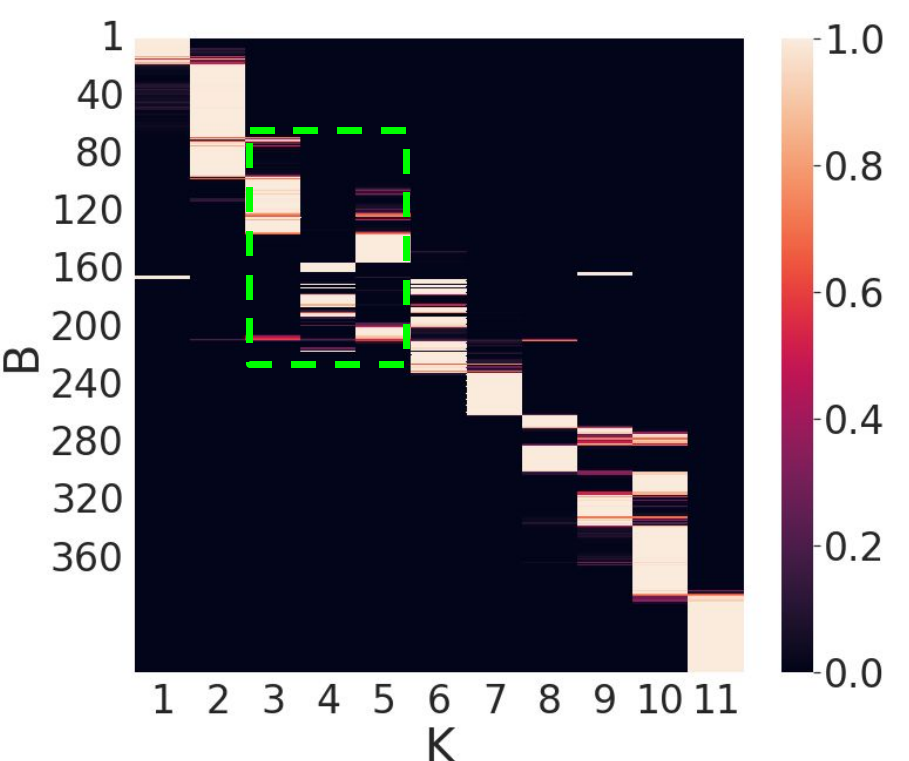}
         \caption{$\boldsymbol{Q}_a$}
         \label{fig:q_mat_a}
     \end{subfigure}
    \caption{(a) Permutation-aware prior distribution $\boldsymbol{M}_{\boldsymbol{T}}$. (b) Alignment-level pseudo-label codes $\boldsymbol{Q}_a$.}
    \label{fig:alignmentlevel}
\end{figure}

Here we describe our frame-level prediction module. In particular, we adopt the joint representation learning and online clustering method of~\cite{kumar2022unsupervised}. Unlike~\cite{kumar2022unsupervised}, we include modules and unsupervised losses in Secs.~\ref{sec:segmentlevel} and~\ref{sec:frametosegment} for exploiting segment-level cues. Also, instead of the MLP encoder of~\cite{kumar2022unsupervised}, we utilize the transformer encoder of~\cite{behrmann2022unified} to capture long-range dependencies via self-attention.

The input frames $\boldsymbol{X}$ are first fed to the transformer encoder $f_{\boldsymbol{\theta}}$ to yield the encoder features $\boldsymbol{E}$. The frame-level predicted codes $\boldsymbol{P}_f$ (with $\boldsymbol{P}^{ij}_f$ denoting the probability that the $i$-th frame in $\boldsymbol{X}$ is assigned to the $j$-th action in $\boldsymbol{A}$) are then computed as $\boldsymbol{P}_f = \textnormal{softmax} \left(\frac{1}{\tau} \boldsymbol{E} \boldsymbol{C}^\top \right)$ with a temperature $\tau$. We follow~\cite{kumar2022unsupervised} to obtain the frame-level pseudo-label codes $\boldsymbol{Q}_f$ by solving the below fixed-order temporal optimal transport problem:
\begin{align}
    \label{eq:tot}
    \max_{\boldsymbol{Q} \in \mathcal{Q}}~~~ Tr(\boldsymbol{Q}^\top \boldsymbol{E} \boldsymbol{C}^\top) - \rho KL(\boldsymbol{Q}||\boldsymbol{M}_{\boldsymbol{A}}),
\end{align}
\begin{align}
    \label{eq:equipartition}
    \mathcal{Q} = \left\{ \boldsymbol{Q}:~~~ \boldsymbol{Q}\boldsymbol{1}_K = \frac{1}{B} \boldsymbol{1}_B, \boldsymbol{Q}^\top \boldsymbol{1}_B = \frac{1}{K} \boldsymbol{1}_K \right\},
\end{align}
where $\rho$ is a balancing parameter, and $\boldsymbol{1}_B$ and $\boldsymbol{1}_K$ are vectors of ones with $B$ and $K$ dimensions respectively. The first term in Eq.~\ref{eq:tot} measures the similarity between the features $\boldsymbol{E}$ and the prototypes $\boldsymbol{C}$, while the second term denotes the Kullback-Leibler divergence between $\boldsymbol{Q}_f$ and the prior distribution $\boldsymbol{M}_{\boldsymbol{A}}$~\cite{su2017order}. In particular, $\boldsymbol{M}_{\boldsymbol{A}}$ assumes the \emph{fixed order} of actions $\boldsymbol{A}$, and enforces initial frames in $\boldsymbol{X}$ to be assigned to initial actions in $\boldsymbol{A}$ and subsequent frames in $\boldsymbol{X}$ to be assigned to subsequent actions in $\boldsymbol{A}$. In Sec.~\ref{sec:segmentlevel}, we will discuss relaxing the above fixed-order prior by introducing the transcript $\boldsymbol{T}$ and enabling permutations of actions. Eq.~\ref{eq:equipartition} represents the \emph{equal partition} constraint, which imposes that each action in $\boldsymbol{A}$ is assigned the same number of frames in $\boldsymbol{X}$ to avoid a trivial solution. As mentioned in~\cite{kumar2022unsupervised}, the method works relatively well for activities with various action lengths since the above equal partition constraint is applied on soft assignments. The solution for the above fixed-order temporal optimal transport problem is:
\begin{align}
    \label{eq:qf}
    \boldsymbol{Q}_{f} = \diag(\boldsymbol{u}) \exp \left( \frac{\boldsymbol{E} \boldsymbol{C}^\top + \rho \log \boldsymbol{M}_{\boldsymbol{A}}}{\rho} \right) \diag(\boldsymbol{v}),
\end{align}
where $\boldsymbol{u} \in \mathbb{R}^{B}$ and $\boldsymbol{v} \in \mathbb{R}^{K}$ are renormalization vectors~\cite{cuturi2013sinkhorn}. Fig.~\ref{fig:framelevel} shows an example of $\boldsymbol{M}_{\boldsymbol{A}}$ and $\boldsymbol{Q}_f$, where the red boxes highlight the fixed order of actions $\{3,4,5\}$. We minimize the below cross-entropy loss with respect to $\boldsymbol{\theta}$ and $\boldsymbol{C}$ (note that we do not backpropagate through $\boldsymbol{Q}_f$):
\begin{align}
    \label{eq:lf}
    L_{f} = - \frac{1}{B} \sum^{B}_{i=1} \sum^{K}_{j=1} \boldsymbol{Q}^{ij}_f \log \boldsymbol{P}^{ij}_f.
\end{align}

\subsection{Unsupervised Segment-Level Prediction}
\label{sec:segmentlevel}

The above module leverages frame-level cues and the fixed-order prior. In this section, we describe the segment-level prediction module to exploit segment-level cues and allow permutations of actions. In particular, we introduce the transcript $\boldsymbol{T}$, which indicates the sequence of actions of $\boldsymbol{A}$ occurring in the input sequence $\boldsymbol{X}$. For example, let us assume $\boldsymbol{A} = [1,2,3,4,5]$, it is possible that $\boldsymbol{T} = [1,3,2,5,4]$, which is a permutation of $\boldsymbol{A}$. We will discuss later how $\boldsymbol{T}$ is estimated for unsupervised training.

Assuming the transcript $\boldsymbol{T}$ is given, we first pass it to the embedding layer $g_{\boldsymbol{\psi}}$ to obtain the transcript features $\boldsymbol{S}$, which are then fed to the transformer decoder $g_{\boldsymbol{\phi}}$. In addition, we also feed the encoder features $\boldsymbol{E}$ (after positional encoding) to the transformer decoder $g_{\boldsymbol{\phi}}$, which performs cross-attention between $\boldsymbol{E}$ and $\boldsymbol{S}$ to yield the decoder features $\boldsymbol{D}$. The segment-level predicted codes $\boldsymbol{P}_s$ (with $\boldsymbol{P}^{ij}_s$ corresponding to the probability that the $i$-th position in $\boldsymbol{T}$ contains the $j$-th action in $\boldsymbol{A}$) are computed by passing the decoder features $\boldsymbol{D}$ to the prediction layer $g_{\boldsymbol{\chi}}$. In practice, we employ the transformer decoder of~\cite{behrmann2022unified}, which computes $\boldsymbol{P}_s$ in an auto-regressive manner, i.e., a part of $\boldsymbol{T}$ up to the $i$-th position is used to predict the ($i$+1)-th row of $\boldsymbol{P}_s$. In parallel, we convert the transcript $\boldsymbol{T}$ into the segment-level pseudo-label codes $\boldsymbol{Q}_s$. Specifically, we set $\boldsymbol{Q}^{ij}_s = 1$ if the $i$-th position in $\boldsymbol{T}$ contains the $j$-th action in $\boldsymbol{A}$, and $\boldsymbol{Q}^{ij}_s = 0$ otherwise. We minimize the following cross-entropy loss between $\boldsymbol{P}_s$ and $\boldsymbol{Q}_s$ with respect to $\boldsymbol{\theta}$, $\boldsymbol{\psi}$, $\boldsymbol{\phi}$, and $\boldsymbol{\chi}$ (note that we do not backpropagate through $\boldsymbol{Q}_s$):
\begin{align}
    \label{eq:ls}
    L_{s} = - \frac{1}{N} \sum^{N}_{i=1} \sum^{K}_{j=1} \boldsymbol{Q}^{ij}_s \log \boldsymbol{P}^{ij}_s.
\end{align}

In contrast with the supervised method of~\cite{behrmann2022unified}, where framewise labels or timestamp labels are required for supervised training, we estimate the transcript $\boldsymbol{T}$ from the frame-level pseudo-label codes $\boldsymbol{Q}_f$ for unsupervised training. For each $j$-th action, we find the $i$-th frame where $\boldsymbol{Q}^{ij}_f$ has the maximum assignment probability along the $j$-th column, yielding an action-frame pair $(j,i)$. Next, we sort all action-frame pairs by their frame indexes. The resulting temporally sorted list of actions is considered as our estimated transcript $\boldsymbol{T}$. Our motivation is that to predict each action correctly, the method only needs to select a single frame correctly, which is easier than obtaining the correct framewise segmentation result. Note that the above imply that $N$ (the length of the transcript $\boldsymbol{T}$) is equal to $K$ (the length of the action list $\boldsymbol{A}$), and our predicted transcript $\boldsymbol{T}$ shares the same set of unique actions with $\boldsymbol{A}$ despite having different orderings. Fig.~\ref{fig:segmentlevel} illustrates an example of computing $\boldsymbol{T}$ from $\boldsymbol{Q}_f$, and computing $\boldsymbol{Q}_s$ from $\boldsymbol{T}$. Similar to~\cite{kumar2022unsupervised}, our method tends to assign a small number of frames to the missing actions, leading to minor impacts on the overall segmentation accuracy. Handling repetitive actions is an interesting topic and remains our future work. As we will show later in Sec.~\ref{sec:sota}, despite using the above simple heuristic for transcript estimation, our method achieves state-of-the-art results on four public datasets.

\subsection{Unsupervised Frame-to-Segment Alignment}
\label{sec:frametosegment}

To further exploit segment-level cues and improve segmentation results, we employ the frame-to-segment alignment module of~\cite{behrmann2022unified}, which matches frame-level features with segment-level features and models permutations of actions. We pass both the encoder features $\boldsymbol{E}$ and the decoder features $\boldsymbol{D}$ (after positional encoding) to the frame-to-segment alignment module, which performs cross-attention between $\boldsymbol{E}$ and $\boldsymbol{D}$ to predicts the alignment-level predicted codes $\boldsymbol{P}_a$. Here, $\boldsymbol{P}^{ij}_a$ corresponds to the probability that the $i$-th frame in $\boldsymbol{X}$ is mapped to the $j$-th action in $\boldsymbol{A}$. We compute $\boldsymbol{P}_a = \textnormal{softmax} \left(\frac{1}{\tau^\prime} \boldsymbol{E} \boldsymbol{D}^\top \right)$ with a temperature $\tau^\prime$. 

Unlike with the supervised method of~\cite{behrmann2022unified}, where framewise labels or timestamp labels are required for supervised training, we propose a modified temporal optimal transport module which is capable of handling permutations of actions to compute the alignment-level pseudo-label codes $\boldsymbol{Q}_a$ for unsupervised training. Specifically, instead of using the prior distribution $\boldsymbol{M}_{\boldsymbol{A}}$ which enforces the fixed order of actions $\boldsymbol{A}$, we utilize the prior distribution $\boldsymbol{M}_{\boldsymbol{T}}$ which imposes the permutation-aware transcript $\boldsymbol{T}$, yielding the permutation-aware temporal optimal transport problem:
\begin{align}
    \label{eq:patot}
    \max_{\boldsymbol{Q} \in \mathcal{Q}}~~~ Tr(\boldsymbol{Q}^\top \boldsymbol{E} \boldsymbol{C}^\top) - \rho KL(\boldsymbol{Q}||\boldsymbol{M}_{\boldsymbol{T}}).
\end{align}
The solution for the permutation-aware temporal optimal transport problem is:
\begin{align}
    \label{eq:qa}
    \boldsymbol{Q}_{a} = \diag(\boldsymbol{u}) \exp \left( \frac{\boldsymbol{E} \boldsymbol{C}^\top + \rho \log \boldsymbol{M}_{\boldsymbol{T}}}{\rho} \right) \diag(\boldsymbol{v}).
\end{align}
Fig.~\ref{fig:alignmentlevel} shows an example of $\boldsymbol{M}_{\boldsymbol{T}}$ and $\boldsymbol{Q}_a$, where the green boxes highlight the permutations of actions $\{3,5,4\}$. This is in contrast with $\boldsymbol{M}_{\boldsymbol{A}}$ and $\boldsymbol{Q}_f$ in Fig.~\ref{fig:framelevel}, where the red boxes highlight the fixed order of actions $\{3,4,5\}$. As we will show later in Sec.~\ref{sec:ablation_pseudolabels}, using the permutation-aware $\boldsymbol{Q}_a$ derived from $\boldsymbol{T}$ yields better performance than using the fixed-order $\boldsymbol{Q}_a$ derived from $\boldsymbol{A}$. We minimize the cross-entropy loss between $\boldsymbol{P}_a$ and $\boldsymbol{Q}_a$ with respect to $\boldsymbol{\theta}$, $\boldsymbol{\psi}$, and $\boldsymbol{\phi}$ (note that we do not backpropagate through $\boldsymbol{Q}_a$):
\begin{align}
    \label{eq:la}
    L_{a} = - \frac{1}{B} \sum^{B}_{i=1} \sum^{K}_{j=1} \boldsymbol{Q}^{ij}_a \log \boldsymbol{P}^{ij}_a.
\end{align}

Our final loss for unsupervised training is a combination of the fixed-order loss $L_f$ (Eq.~\ref{eq:lf}) and the permutation-aware losses $L_s$ (Eq.~\ref{eq:ls}) and $L_a$ (Eq.~\ref{eq:la}):
\begin{align}
    \label{eq:lfinal}
    L = L_{f} + \alpha L_{s} + \beta L_{a},
\end{align}
where $\alpha$ and $\beta$ are the balancing parameters for $L_{s}$ and $L_{a}$ respectively. Following~\cite{behrmann2022unified}, we set $\alpha = \beta = 1$.

\section{Experiments}
\label{sec:experiments}

\noindent \textbf{Implementation Details.}
We train our model in two stages. In the first stage, we train only the frame-level prediction module with the loss in Eq.~\ref{eq:lf} for 30 epochs, which is then used for initialization in the second stage, where we train the entire model with the loss in Eq.~\ref{eq:lfinal} for 70 epochs. Note that we reduce the transformer encoder and transformer decoder of~\cite{behrmann2022unified} to two layers to avoid overfitting. We implement our approach in pyTorch~\cite{paszke2017automatic}. We use ADAM optimization~\cite{kingma2014adam} with a learning rate of $10^{-3}$ and a weight decay  of $10^{-5}$. For inference, we follow~\cite{kumar2022unsupervised} to compute cluster assignment probabilities for all frames and then pass them to a Viterbi decoder which smooths out the probabilities given the action order $\boldsymbol{T}$ (instead of $\boldsymbol{A}$ in~\cite{kumar2022unsupervised}). More details are provided in the supplementary material.

\noindent \textbf{Competing Methods.} We compare our approach, namely \emph{UFSA} (short for \emph{U}nsupervised \emph{F}rame-to-\emph{S}egment \emph{A}lignment), against a narration-based method~\cite{alayrac2016unsupervised}, sequential learning and clustering methods~\cite{sener2018unsupervised,kukleva2019unsupervised,vidalmata2021joint,li2021action,wang2022sscap}, and joint learning and clustering methods~\cite{swetha2021unsupervised,kumar2022unsupervised}.

\noindent \textbf{Datasets.}
We evaluate our approach on four public datasets, i.e., 50 Salads~\cite{stein2013combining}, YouTube Instructions (YTI)~\cite{alayrac2016unsupervised}, Breakfast~\cite{kuehne2014language}, and Desktop Assembly~\cite{kumar2022unsupervised}:
\begin{itemize}
    \item \emph{50 Salads} includes 50 videos capturing 25 actors making 2 types of salads. The total duration of all videos is over 4.5 hours with an average of 10k frames per video. We test on 2 granularity levels, i.e., \emph{Eval} with 12 action classes and \emph{Mid} with 19 action classes. Following~\cite{kukleva2019unsupervised}, we use pre-computed features by~\cite{wang2013action}.
    \item \emph{YouTube Instructions (YTI)} contains 150 videos capturing 5 activities with 47 action classes in total and an average video length of about 2 minutes. These videos contain many background frames. We use pre-computed features provided by~\cite{alayrac2016unsupervised}.
    \item \emph{Breakfast} includes 70 hours of videos (30 seconds to a few minutes long per video) capturing 10 cooking activities with 48 action classes in total. We follow~\cite{sener2018unsupervised} to use pre-computed features proposed by~\cite{kuehne2016end}.
    \item \emph{Desktop Assembly} contains 2 sets of videos. \emph{Orig} contains 76 videos of 4 actors performing desktop assembly in a fixed order. \emph{Extra} includes all \emph{Orig} videos and additionally 52 videos with permuted and missing steps, yielding 128 videos in total. We evaluate on both sets using pre-computed features provided by~\cite{kumar2022unsupervised}.
\end{itemize}

\noindent \textbf{Evaluation Metrics.} Following~\cite{sener2018unsupervised,kukleva2019unsupervised,kumar2022unsupervised}, we perform Hungarian matching between ground truth and predicted segments, which is conducted at the activity level. This is unlike the Hungarian matching performed at the video level in~\cite{aakur2019perceptual,sarfraz2021temporally,du2022fast}. Note that video-level segmentation, e.g., ABD~\cite{du2022fast}, (i.e., segmenting just a single video) is a sub-problem and in general easier than activity-level segmentation, e.g., our work, (i.e., jointly segmenting and clustering frames across all videos). Due to space limits, we convert video-level segmentation results of ABD~\cite{du2022fast} to activity-level segmentation results via K-Means and evaluate them in the supplementary material. We compute Mean Over Frames (MOF), i.e., the percentage of frames with correct predictions averaged over all activities, and F1-Score, i.e., the harmonic mean of precision and recall, where only positive detections with more than 50\% overlap with ground truth segments are considered. We compute F1-Score for each video and take the average over all videos. 

\subsection{Ablation Studies}

\subsubsection{Impacts of Different Model Components}
\label{sec:ablation_components}
We first study the effects of various network components on the 50 Salads (\emph{Eval} granularity) and YTI datasets. The results are reported in Tab.~\ref{tab:ablation_components}. Firstly, using only the frame-level prediction module presented in Sec.~\ref{sec:framelevel} yields the lowest overall results. The frame-level prediction module exploits frame-level cues only and utilizes the fixed-order prior which does not account for permutations of actions. Next, we expand the network by adding the segment-level prediction module described in Sec.~\ref{sec:segmentlevel} to exploit segment-level cues. For 50 Salads, MOF is not changed much, while F1-score is improved by 3.7\%. For YTI, MOF is increased by 2.2\%,  while F1-Score is slightly improved by 0.6\%. Although the segment-level prediction module estimates the permutation-aware transcript, the framewise predictions are still suffered from over-segmentation. To address that, the frame-to-segment alignment module proposed in Sec.~\ref{sec:frametosegment} is appended to the network to simultaneously leverage frame-level cues and segment-level cues and refine the framewise predictions, leading to significant performance gains. On 50 Salads, the results are boosted to 55.8\% and 50.3\% for MOF and F1-Score respectively, while on YTI, MOF is increased to 49.6\% and F1-Score to 32.4\%.
\begin{table}[t]

    \centering
    \footnotesize
    \begin{tabular}{||c|c|c|c||}
    \hline
    & \textbf{Method} &\textbf{MOF} &\textbf{F1}\\
    \hline 
    \multirow{3}{*}{\rotatebox[origin=c]{90}{\textbf{\footnotesize{Eval}}}}
    & Frame & 43.1 & 34.4 \\
    & Frame+Segment & \underline{\textit{43.2}} & \underline{\textit{38.1}} \\
    & \cellcolor{babyblueeyes} Frame+Segment+Alignment & \cellcolor{babyblueeyes} \textbf{55.8} & \cellcolor{babyblueeyes} \textbf{50.3} \\
    \hline
    \multirow{3}{*}{\rotatebox[origin=c]{90}{\textbf{\footnotesize{YTI}}}}
    & Frame & 42.8 &  30.2 \\
    & Frame+Segment &  \underline{\textit{45.0}} & \underline{\textit{30.8}} \\
    & \cellcolor{babyblueeyes} Frame+Segment+Alignment & \cellcolor{babyblueeyes} \textbf{49.6} & \cellcolor{babyblueeyes} \textbf{32.4} \\
    \hline
    \end{tabular}

    \caption{Impacts of different model components on 50 Salads with the Eval granularity (\emph{Eval}) and YouTube Instructions (\emph{YTI}). Best results are in \textbf{bold}, while second best ones are \underline{\textit{underlined}}.}
    \label{tab:ablation_components}
\end{table}

\subsubsection{Impacts of Different Pseudo Labels}
\label{sec:ablation_pseudolabels}
Here, we conduct an ablation study on the 50 Salads (\emph{Eval} granularity) and YTI datasets by using various versions of pseudo labels $\boldsymbol{Q}_s$ and $\boldsymbol{Q}_a$ computed from either the fixed order of actions $\boldsymbol{A}$ or the permutation-aware transcript $\boldsymbol{T}$. Tab.~\ref{tab:ablation_labels} presents the results.  Firstly, using the fixed order of actions $\boldsymbol{A}$ for computing both $\boldsymbol{Q}_s$ and $\boldsymbol{Q}_a$ (i.e., we use $\boldsymbol{T} = \boldsymbol{A}$ in both Secs.~\ref{sec:segmentlevel} and~\ref{sec:frametosegment}) yields the lowest overall numbers on both datasets, i.e., on 50 Salads, 46.1\% and 45.2\% for MOF and F1-Score respectively, and on YTI, 44.3\% and 29.4\% for MOF and F1-Score respectively. Next, we experiment with using the permutation-aware transcript $\boldsymbol{T}$ for computing either $\boldsymbol{Q}_s$ or $\boldsymbol{Q}_a$, resulting in performance gains, e.g., for the former ($\boldsymbol{T}$ for $\boldsymbol{Q}_s$, $\boldsymbol{A}$ for $\boldsymbol{Q}_a$), we achieve 50.8\% for MOF and 46.9\% for F1-Score on 50 Salads, while for the latter ($\boldsymbol{A}$ for $\boldsymbol{Q}_s$, $\boldsymbol{T}$ for $\boldsymbol{Q}_a$), we obtain 54.0\% for MOF and 48.7\% for F1-Score on 50 Salads. Finally, we employ the permutation-aware transcript $\boldsymbol{T}$ for computing both $\boldsymbol{Q}_s$ and $\boldsymbol{Q}_a$, leading to the best performance on both datasets, i.e., 55.8\% for MOF and 50.3\% for F1-Score on 50 Salads, and 49.6\% for MOF and 32.4\% for F1-Score on YTI. The above results confirm the benefits of using the permutation-aware transcript $\boldsymbol{T}$ for computing both pseudo labels $\boldsymbol{Q}_s$ and $\boldsymbol{Q}_a$.

\begin{table}[t]

    \centering
    \footnotesize
    \begin{tabular}{||c|c|c|c|c||}
    \hline
    & \textbf{$\boldsymbol{Q}_s$} & \textbf{$\boldsymbol{Q}_a$} &\textbf{MOF}  &\textbf{F1}\\
    \hline 
    \multirow{4}{*}{\rotatebox[origin=c]{90}{\textbf{\footnotesize{Eval}}}}
    & $\boldsymbol{A}$ & $\boldsymbol{A}$ & 46.1 & 45.2 \\
    & $\boldsymbol{T}$ & $\boldsymbol{A}$ &50.8 & 46.9  \\
    & $\boldsymbol{A}$ & $\boldsymbol{T}$ & \underline{\textit{54.0}} & \underline{\textit{48.7}} \\
    & \cellcolor{babyblueeyes}$\boldsymbol{T}$ & \cellcolor{babyblueeyes} $\boldsymbol{T}$ & \cellcolor{babyblueeyes} \textbf{ 55.8} & \cellcolor{babyblueeyes} \textbf{50.3} \\
    \hline
    \multirow{4}{*}{\rotatebox[origin=c]{90}{\textbf{\footnotesize{YTI}}}}
    & $\boldsymbol{A}$ & $\boldsymbol{A}$ & 44.3 & 29.4 \\
    & $\boldsymbol{T}$ & $\boldsymbol{A}$ & 45.7 & 29.7  \\
    & $\boldsymbol{A}$ & $\boldsymbol{T}$ & \underline{\textit{46.5}} & \underline{\textit{29.8}} \\
    & \cellcolor{babyblueeyes}$\boldsymbol{T}$ & \cellcolor{babyblueeyes} $\boldsymbol{T}$ & \cellcolor{babyblueeyes} \textbf{49.6} & \cellcolor{babyblueeyes} \textbf{32.4} \\
    \hline
    \end{tabular}

    \caption{Impacts of different pseudo labels on 50 Salads with the Eval granularity (\emph{Eval}) and YouTube Instructions (\emph{YTI}). Best results are in \textbf{bold}, while second best ones are \underline{\textit{underlined}}.}
    \label{tab:ablation_labels}
\end{table}

\subsection{Comparisons with the State-of-the-Art}
\label{sec:sota}

\subsubsection{Results on 50 Salads}
We now compare the performance of our approach with state-of-the-art unsupervised activity segmentation methods on the 50 Salads dataset for both granularities, i.e., \emph{Eval} and \emph{Mid}. Tab.~\ref{tab:50salads_results} illustrates the results. It is evident from Tab.~\ref{tab:50salads_results} that our approach obtains the best MOF and F1-Score numbers on both granularities, outperforming all competing methods. In particular, UFSA outperforms TOT~\cite{kumar2022unsupervised} by 8.4\% and 4.9\% on MOF on the \emph{Eval} and \emph{Mid} granularities respectively, and UDE~\cite{swetha2021unsupervised} by 15.9\% on F1-Score on the \emph{Eval} granularity. Although TOT~\cite{kumar2022unsupervised} and UDE~\cite{swetha2021unsupervised} conduct joint representation learning and online clustering as our approach, they only exploit frame-level cues, whereas UFSA leverages segment-level cues as well. Moreover, our approach achieves better results than SSCAP~\cite{wang2022sscap}, which uses recent self-supervised learning features~\cite{epstein2020oops}, and ASAL~\cite{li2021action}, which exploits segment-level cues via action shuffling, e.g., on the \emph{Eval} granularity, UFSA achieves 55.8\% MOF, whereas SSCAP~\cite{wang2022sscap} and ASAL~\cite{li2021action} obtain 41.4\% MOF and 39.2\% MOF respectively. The substantial improvements of UFSA over previous methods demonstrate the effectiveness of our approach.
\begin{table}[t]

    \centering
    \footnotesize
    \begin{tabular}{||c| c| c| c| c||}
    \hline
    \multicolumn{1}{||c|}{\textbf{Method}} & \multicolumn{2}{c|}{\textbf{Eval}} & \multicolumn{2}{c||}{\textbf{Mid}}\\
\cline{2-5}
     & \textbf{MOF} & \textbf{F1} & \textbf{MOF} & \textbf{F1} \\

     \hline    
        CTE~\cite{kukleva2019unsupervised}
         &35.5&36.3 &30.2&25.6\\
         VTE~\cite{vidalmata2021joint} 
         & 30.6&- &24.2&- \\
         ASAL~\cite{li2021action} &39.2 &- &\underline{\textit{34.4}} &- \\
         UDE~\cite{swetha2021unsupervised} & 42.2&34.4& -&- \\
         SSCAP~\cite{wang2022sscap} &41.4&30.3& -&  - \\
         TOT~\cite{kumar2022unsupervised} &\underline{\textit{47.4}}&42.8 &31.8 
         &22.5 \\
         TOT+TCL~\cite{kumar2022unsupervised} &44.5& \underline{\textit{48.2}} &34.3 &\underline{\textit{28.9}} \\
         
         \rowcolor{babyblueeyes} Ours (UFSA) & \textbf{ 55.8} &\textbf{50.3} & \textbf{36.7} & \textbf{30.4} \\
    \hline
    
    \end{tabular}

    \caption{Results on 50 Salads. \emph{Eval} denotes the Eval granularity, while \emph{Mid} denotes the Mid granularity. Best results are in \textbf{bold}, while second best ones are \underline{\textit{underlined}}.}
    \label{tab:50salads_results}
\end{table}

\subsubsection{Results on YouTube Instructions}
\begin{figure}[t]
    \centering
	\includegraphics[width=0.92\linewidth, trim = 0mm 5mm 0mm 0mm, clip]{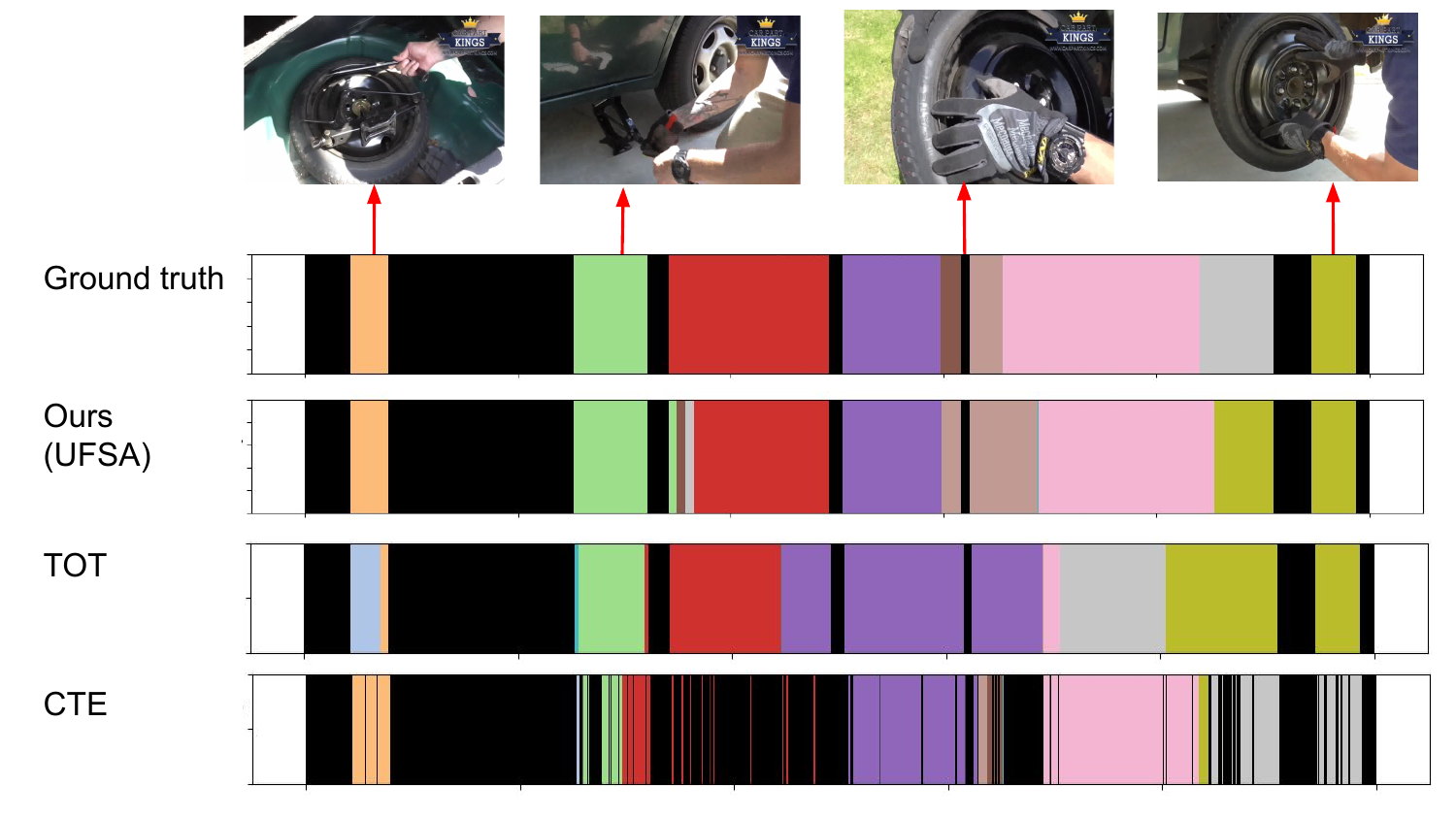}
    \caption{Segmentation results on a YouTube Instructions video (\emph{changing\_tire\_0005}). Black color indicates background frames.}
    \label{fig:yti_exmaple}
\end{figure}

Tab.~\ref{tab:yti_results} presents the quantitative results of our approach along with previous unsupervised activity segmentation methods on the YTI dataset. We follow the protocol of prior works and report the accuracy excluding the background frames. It is clear from Tab.~\ref{tab:yti_results} that our approach achieves the best MOF, outperforming all previous methods, and the second best F1-Score, slightly worse than TOT+TCL~\cite{kumar2022unsupervised} (note that our approach currently relies on TOT only, and can further include TCL for potential improvements). Specifically, UFSA has an improvement of 9.0\% MOF and 2.4\% F1-Score over TOT~\cite{kumar2022unsupervised}, and an improvement of 5.8\% MOF and 2.8\% F1-Score over UDE~\cite{swetha2021unsupervised}. In addition, our approach obtains a noticeable gain of 4.7\% MOF and a slight gain of 0.3\% F1-Score over ASAL~\cite{li2021action}. Fig.~\ref{fig:yti_exmaple} plots the qualitative results of UFSA, TOT~\cite{kumar2022unsupervised}, and CTE~\cite{kukleva2019unsupervised} on a YTI video. Our approach demonstrates significant advantages over CTE~\cite{kukleva2019unsupervised} and TOT~\cite{kumar2022unsupervised} in terms of capturing the temporal order of actions and aligning them closely with the ground truth. Due to space constraints, please refer to the supplementary material for more qualitative examples, especially with permuted, missing, and repetitive actions.
\begin{table}[t]

    \centering
    \footnotesize
    \begin{tabular}{||c| c| c||}
    \hline
         \textbf{Method} &\textbf{MOF}  &\textbf{F1}\\
     \hline 
         Frank-Wolfe~\cite{alayrac2016unsupervised} & - & 24.4 \\   
         Mallow~\cite{sener2018unsupervised} & 27.8&27.0 \\   
         CTE~\cite{kukleva2019unsupervised}
         &39.0 &28.3\\
         VTE~\cite{vidalmata2021joint} 
         & - & 29.9 \\
         ASAL~\cite{li2021action} &44.9 &32.1 \\
         UDE~\cite{swetha2021unsupervised} & 43.8& 29.6 \\
         TOT~\cite{kumar2022unsupervised} &40.6 &30.0 \\
         TOT+TCL~\cite{kumar2022unsupervised} & \underline{\textit{45.3}} & \textbf{32.9} \\
         \rowcolor{babyblueeyes} Ours (UFSA) & \textbf{ 49.6} & \underline{\textit{32.4}} \\
    \hline
    
    \end{tabular}

    \caption{Results on YouTube Instructions. Best results are in \textbf{bold}, while second best ones are \underline{\textit{underlined}}.}
    \label{tab:yti_results}
\end{table}

\subsubsection{Results on Breakfast}
Tab.~\ref{tab:breakfast_results} includes the performance of different methods on the Breakfast dataset. From Tab.~\ref{tab:breakfast_results}, our results are on par with ASAL~\cite{li2021action}, which leverages segment-level information via action shuffling, and SSCAP~\cite{wang2022sscap}, which employs more sophisticated self-supervised features~\cite{epstein2020oops}. Particularly, ASAL~\cite{li2021action} and SSCAP~\cite{wang2022sscap} yield the best MOF number (i.e., 52.5\%) and the best F1-Score number (i.e., 39.2\%) respectively, while UFSA achieves the second best results for both metrics (i.e., 52.1\% and 38.0\%). In addition, our approach outperforms a number of competing methods, namely Mallow~\cite{sener2018unsupervised}, CTE~\cite{kukleva2019unsupervised}, VTE~\cite{vidalmata2021joint}, UDE~\cite{swetha2021unsupervised}, and TOT~\cite{kumar2022unsupervised}, which exploit frame-level cues only. 
\begin{table}[t]

    \centering
    \footnotesize
    \begin{tabular}{||c| c| c||}
    \hline
        \textbf{Method}  &\textbf{MOF}  &\textbf{F1}\\
     \hline 
            Mallow~\cite{sener2018unsupervised} & 34.6&- \\
            
         CTE~\cite{kukleva2019unsupervised}
         &41.8 &26.4\\
         VTE~\cite{vidalmata2021joint} 
         & 48.1 & -\\
         ASAL~\cite{li2021action} &\textbf{52.5} &37.9 \\
         UDE~\cite{swetha2021unsupervised} & 47.4& 31.9 \\
         SSCAP~\cite{wang2022sscap} & 51.1 & \textbf{39.2} \\
         TOT~\cite{kumar2022unsupervised} &47.5 &31.0 \\
         TOT+TCL~\cite{kumar2022unsupervised} &39.0 &30.3 \\
         \rowcolor{babyblueeyes} Ours (UFSA) & \underline{\textit{52.1}} & \underline{\textit{38.0}} \\
    \hline
    
    \end{tabular}

    \caption{Results on Breakfast. Best results are in \textbf{bold}, while second best ones are \underline{\textit{underlined}}.}
    \label{tab:breakfast_results}
\end{table}

\subsubsection{Results on Desktop Assembly}
We test the performance of our approach on the Desktop Assembly dataset for both \emph{Orig} and \emph{Extra} sets. The results are reported in Tab.~\ref{tab:desktop_assembly_results}, which shows superior performance of our approach over CTE~\cite{kukleva2019unsupervised}, TOT~\cite{kumar2022unsupervised}, and TOT+TCL~\cite{kumar2022unsupervised}. For example, UFSA achieves an improvement of 14.9\% MOF and 20.5\% F1-Score over TOT~\cite{kumar2022unsupervised} on the \emph{Orig} set, and a gain of 7.6\% MOF and 15.5\% F1-Score over TOT~\cite{kumar2022unsupervised} on the \emph{Extra} set. Results on the \emph{Orig} set indicate the effectiveness of our approach in preserving the fixed order of actions, while results on the \emph{Extra} set show the ability of our method in handling permuted actions. 
\begin{table}[t]

    \centering
    \footnotesize
    \begin{tabular}{||c|c|c|c||}
    \hline
    & \textbf{Method} &\textbf{MOF}  &\textbf{F1}\\
    \hline 
    \multirow{3}{*}{\rotatebox[origin=c]{90}{\textbf{\footnotesize{Orig}}}}
    & CTE~\cite{kukleva2019unsupervised} & 47.6 & 44.9\\
    & TOT~\cite{kumar2022unsupervised} & 56.3& 51.7\\
    & TOT+TCL~\cite{kumar2022unsupervised} & \underline{\textit{58.1}} & \underline{\textit{53.4}}\\
    & \cellcolor{babyblueeyes} Ours (UFSA) & \cellcolor{babyblueeyes} \textbf{65.4} & \cellcolor{babyblueeyes} \textbf{63.0} \\
    \hline
    \multirow{3}{*}{\rotatebox[origin=c]{90}{\textbf{\footnotesize{Extra}}}}
    & CTE~\cite{kukleva2019unsupervised} & 40.8 & 35.6 \\
    & TOT~\cite{kumar2022unsupervised} & 51.0 & 40.4\\
    & TOT+TCL~\cite{kumar2022unsupervised} & \underline{\textit{57.9}} & \underline{\textit{54.0}}\\
    & \cellcolor{babyblueeyes} Ours (UFSA) & \cellcolor{babyblueeyes} \textbf{58.6} & \cellcolor{babyblueeyes} \textbf{55.9} \\
    \hline
    \end{tabular}
    
    \caption{Results on Desktop Assembly. \emph{Orig} includes original fixed-order videos only, while \emph{Extra} further includes additional permuted-step and missing-step videos. Best results are in \textbf{bold}, while second best ones are \underline{\textit{underlined}}.}
    \label{tab:desktop_assembly_results}
\end{table}

\subsubsection{Generalization Results}
We follow~\cite{kumar2022unsupervised} to evaluate the generalization ability of our approach. We divide the datasets, i.e., 50 Salads (\emph{Eval}), YTI, Breakfast, Desktop Assembly (\emph{Orig}, \emph{Extra}) into 80\% for training and 20\% for testing. For instance, for 50 Salads with 50 videos, 40 videos are used for training and 10 for testing. Tab.~\ref{tab:generalization_results} shows the results. UFSA continues to outperform CTE~\cite{kukleva2019unsupervised}, TOT~\cite{kumar2022unsupervised}, and TOT+TCL~\cite{kumar2022unsupervised} in this experiment setting. Note the results of CTE~\cite{kukleva2019unsupervised}, TOT~\cite{kumar2022unsupervised}, and TOT+TCL~\cite{kumar2022unsupervised} in Tab.~\ref{tab:generalization_results} differ from those reported in~\cite{kumar2022unsupervised} since different training/testing splits are used (we could not acquire the splits from the authors of~\cite{kumar2022unsupervised}). Our splits are available at \url{https://tinyurl.com/57ya6653}.
\begin{table}[t]

    \centering
    \footnotesize
    \begin{tabular}{||c|c|c|c||}
    \hline
    & \textbf{Method} &\textbf{MOF}  &\textbf{F1}\\
    \hline
    \multirow{3}{*}{\rotatebox[origin=c]{90}{\textbf{\footnotesize{Eval}}}}
    & CTE~\cite{kukleva2019unsupervised} & 28.6 & 26.4 \\
    & TOT~\cite{kumar2022unsupervised} & 39.8 & 37.0 \\
    & TOT+TCL~\cite{kumar2022unsupervised} & \underline{\textit{42.8}} & \textbf{44.9} \\
    & \cellcolor{babyblueeyes} Ours (UFSA) & \cellcolor{babyblueeyes} \textbf{47.6} & \cellcolor{babyblueeyes} \underline{\textit{41.8}} \\
    \hline
    \multirow{3}{*}{\rotatebox[origin=c]{90}{\textbf{\footnotesize{YTI}}}}
    & CTE~\cite{kukleva2019unsupervised} & 38.4 & 25.5 \\
    & TOT~\cite{kumar2022unsupervised} & 40.4 & \underline{\textit{28.0}} \\
    & TOT+TCL~\cite{kumar2022unsupervised} & \underline{\textit{40.6}} & 26.7 \\
    & \cellcolor{babyblueeyes} Ours (UFSA) & \cellcolor{babyblueeyes} \textbf{46.8} & \cellcolor{babyblueeyes} \textbf{28.2} \\
    \hline
    \multirow{3}{*}{\rotatebox[origin=c]{90}{\textbf{\footnotesize{Breakfast}}}}
    & CTE~\cite{kukleva2019unsupervised} & 39.8 & 25.5 \\
    & TOT~\cite{kumar2022unsupervised} & \underline{\textit{40.6}} & \underline{\textit{27.6}} \\
    & TOT+TCL~\cite{kumar2022unsupervised} & 37.4 & 23.2 \\
    & \cellcolor{babyblueeyes} Ours (UFSA) & \cellcolor{babyblueeyes} \textbf{44.0}  & \cellcolor{babyblueeyes} \textbf{36.7} \\
    \hline
    \multirow{3}{*}{\rotatebox[origin=c]{90}{\textbf{\footnotesize{Orig}}}}
    & CTE~\cite{kukleva2019unsupervised} & 35.6 & 31.8 \\
    & TOT~\cite{kumar2022unsupervised} & \underline{\textit{55.3}} & \underline{\textit{50.2}} \\
    & TOT+TCL~\cite{kumar2022unsupervised} & 49.2 & 44.6 \\
    & \cellcolor{babyblueeyes} Ours (UFSA) & \cellcolor{babyblueeyes} \textbf{63.9}  & \cellcolor{babyblueeyes} \textbf{63.7} \\
    \hline
    \multirow{3}{*}{\rotatebox[origin=c]{90}{\textbf{\footnotesize{Extra}}}}
    & CTE~\cite{kukleva2019unsupervised} &35.7  &30.4  \\
    & TOT~\cite{kumar2022unsupervised} &  43.6& 35.0 \\
    & TOT+TCL~\cite{kumar2022unsupervised} & \underline{\textit{45.9}} & \underline{\textit{40.0}} \\
    & \cellcolor{babyblueeyes} Ours (UFSA) &\cellcolor{babyblueeyes} \textbf{57.9}  & \cellcolor{babyblueeyes}  \textbf{54.0} \\
    \hline
    \end{tabular}
    
    \caption{Generalization results. Best results are in \textbf{bold}, while second best ones are underlined}.
    \label{tab:generalization_results}
\end{table}

\section{Conclusion}
\label{sec:conclusion}

We propose a novel combination of modules and unsupervised losses to exploit both frame-level cues and segment-level cues for permutation-aware activity segmentation. Our approach includes a frame-level prediction module which uses a transformer encoder for obtaining framewise action classes and is trained in unsupervised manner via temporal optimal transport. To leverage segment-level cues, we utilize a segment-level prediction model based on a transformer decoder for predicting video transcripts and a frame-to-segment alignment module for corresponding frame-level features with segment-level features, resulting in permutation-aware segmentation results. For unsupervised training of the above modules, we introduce simple-yet-effective pseudo labels. We show comparable or superior results over prior methods on four public datasets.

\appendix
\section{Supplementary Material}
This supplementary material begins with showing some qualitative results in Sec.~\ref{sec:qualitative}. Next, we present the ablation results of using MLP encoder and using $\boldsymbol{A}$ in segment-/alignment-level modules in Secs.~\ref{sec:supp_mlp} and \ref{sec:supp_a} respectively, and adopt the video-level segmentation method of ABD~\cite{du2022fast} for the activity-level segmentation task in Sec.~\ref{sec:abd}. Finally, Sec.~\ref{sec:implementation} provides the details of our implementation, while Sec.~\ref{sec:societal} includes a discussion on the societal impacts of our work.

\subsection{Qualitative Results}
\label{sec:qualitative}

\begin{figure}[ht!]
     \centering
     \begin{subfigure}[b]{0.49\textwidth}
         \centering
         \includegraphics[width=\textwidth, trim = 5mm 5mm 5mm 0mm]{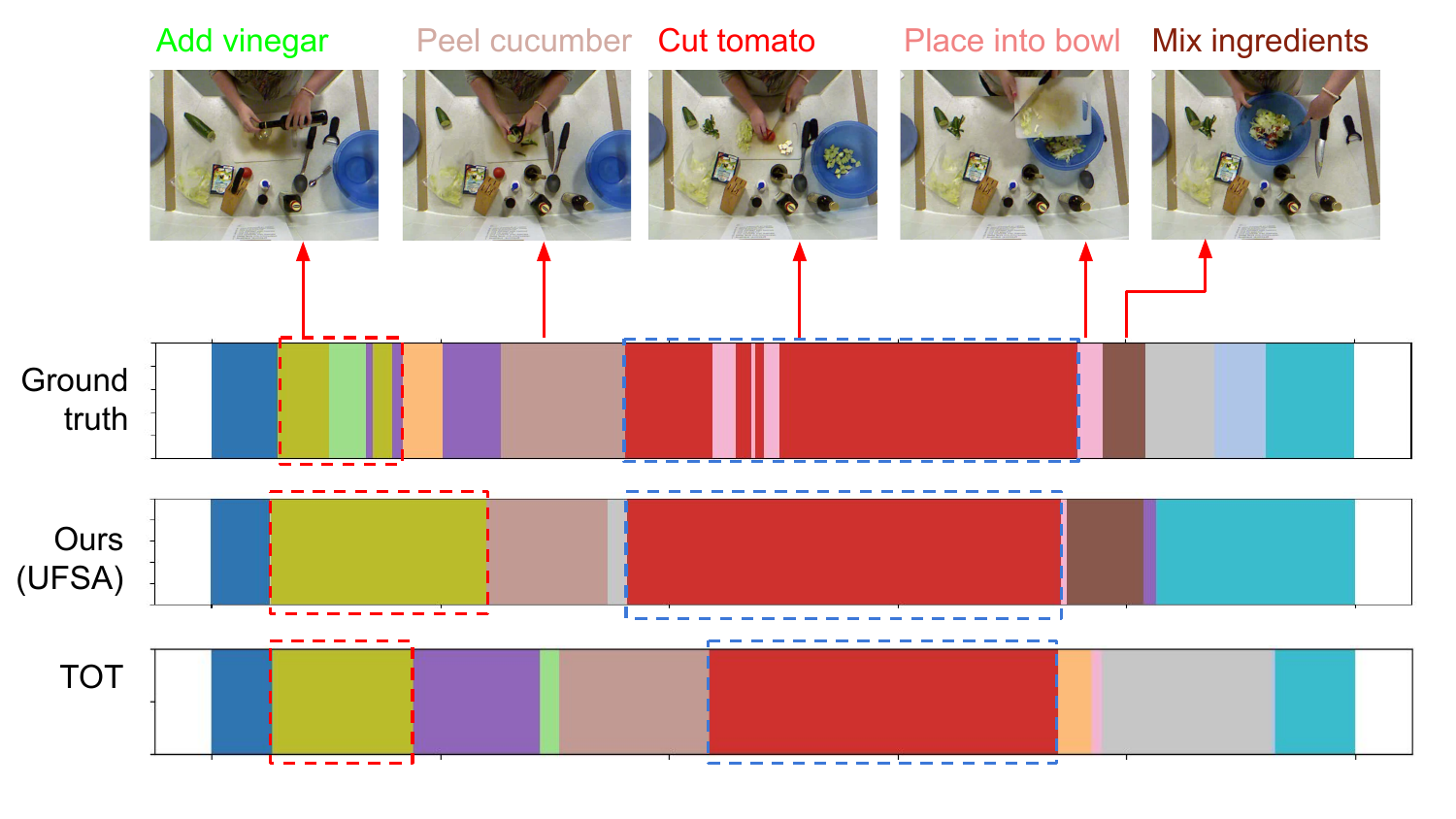}
         \caption{50 Salads (\emph{rgb-21-01}).}
         \label{fig:fs_perm_1}
     \end{subfigure}
     \hfill
     \centering
     \begin{subfigure}[b]{0.49\textwidth}
         \centering
         \includegraphics[width=\textwidth, trim = 5mm 5mm 5mm 0mm]{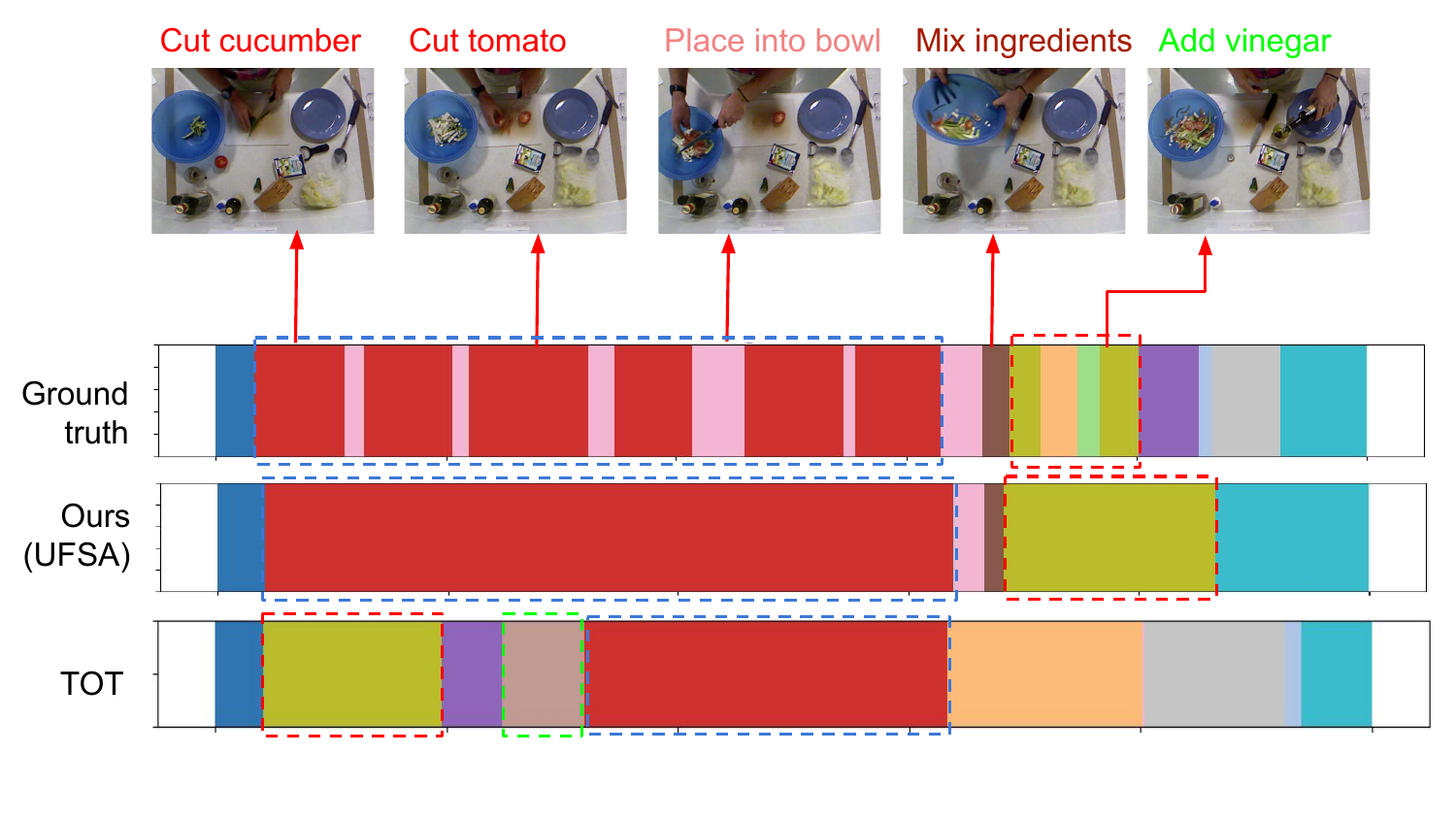}
         \caption{50 Salads (\emph{rgb-15-02}).}
         \label{fig:fs_perm_2}
     \end{subfigure}
    
     \caption{Segmentation results on two 50 Salads videos (\emph{Eval} granularity). Red boxes highlight permuted actions. Green boxes highlight missing actions. Blue boxes highlight repetitive actions.}
     \label{fig:qual_perm}
\end{figure}

Fig.~\ref{fig:qual_perm} illustrates the segmentation results of our approach and TOT~\cite{kumar2022unsupervised} on two 50 Salads videos (\emph{Eval} granularity). From Fig.~\ref{fig:qual_perm}, UFSA shows superior performance in extracting the permutation of actions. For example, let us consider the \emph{`Add vinegar'} action (highlighted by red boxes) which happens at different temporal positions in the videos, UFSA captures the permutation of actions correctly, while TOT~\cite{kumar2022unsupervised} maintains the fixed order of actions and hence fails to recognize the permutation of actions. Next, for actions that are missing, such as the \emph{`Peel cucumber'} action, which occurs in Fig.~\ref{fig:fs_perm_1} but does not appear in Fig.~\ref{fig:fs_perm_2}, UFSA associates a negligible number of frames with this action class, whereas TOT~\cite{kumar2022unsupervised} incorrectly assigns a large number of frames (highlighted by a green box). 

\begin{figure}[ht!]
     \centering
     \begin{subfigure}[b]{0.49\textwidth}
         \centering
         \includegraphics[width=\textwidth, trim = 5mm 0mm 0mm 0mm]{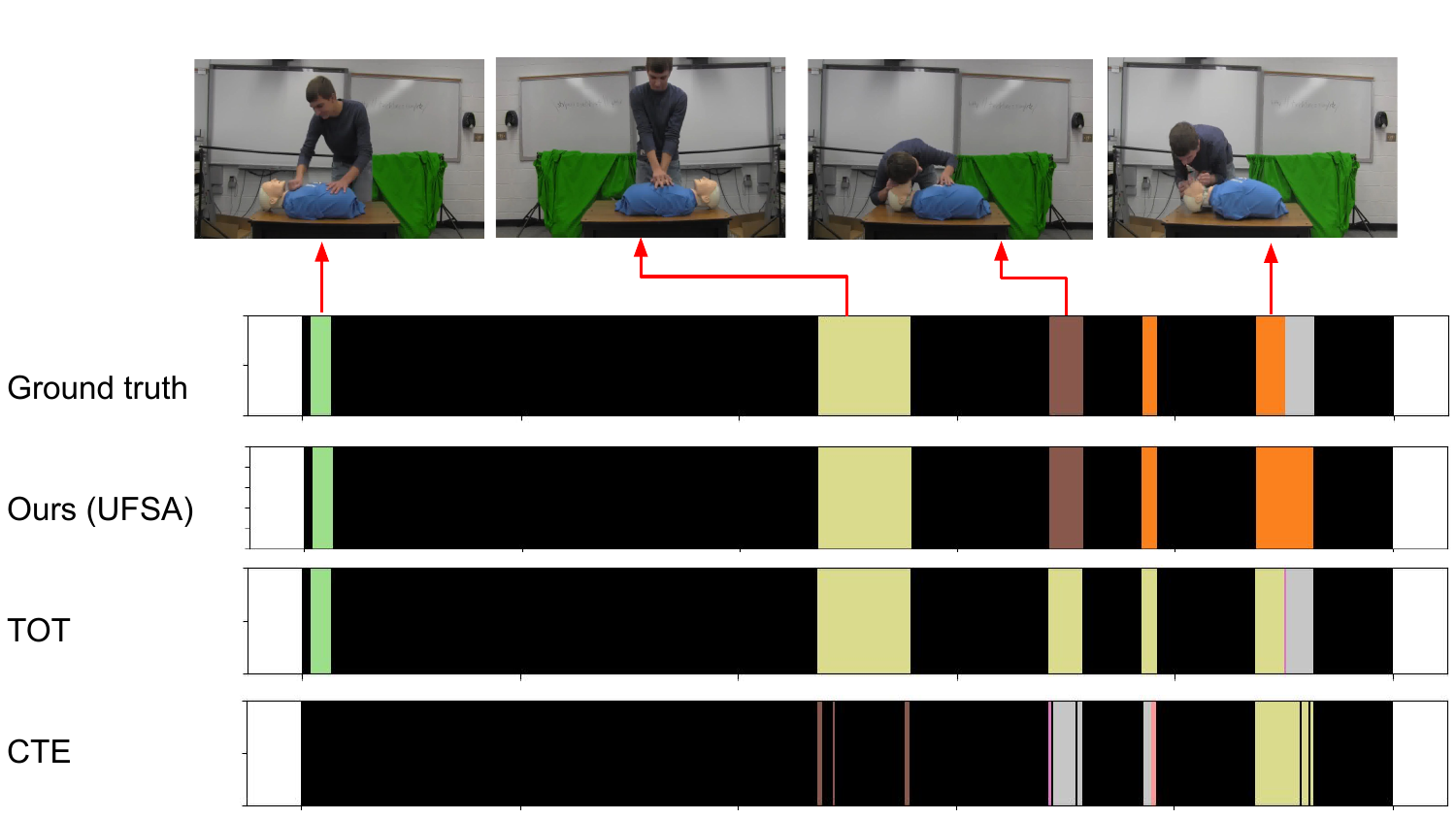}
         \caption{YouTube Instructions (\emph{cpr\_0027}).}
         \label{fig:qual_yti}
     \end{subfigure}
     \hfill
     \centering
     \begin{subfigure}[b]{0.49\textwidth}
         \centering
         \includegraphics[width=\textwidth, trim = 5mm 0mm 0mm 0mm]{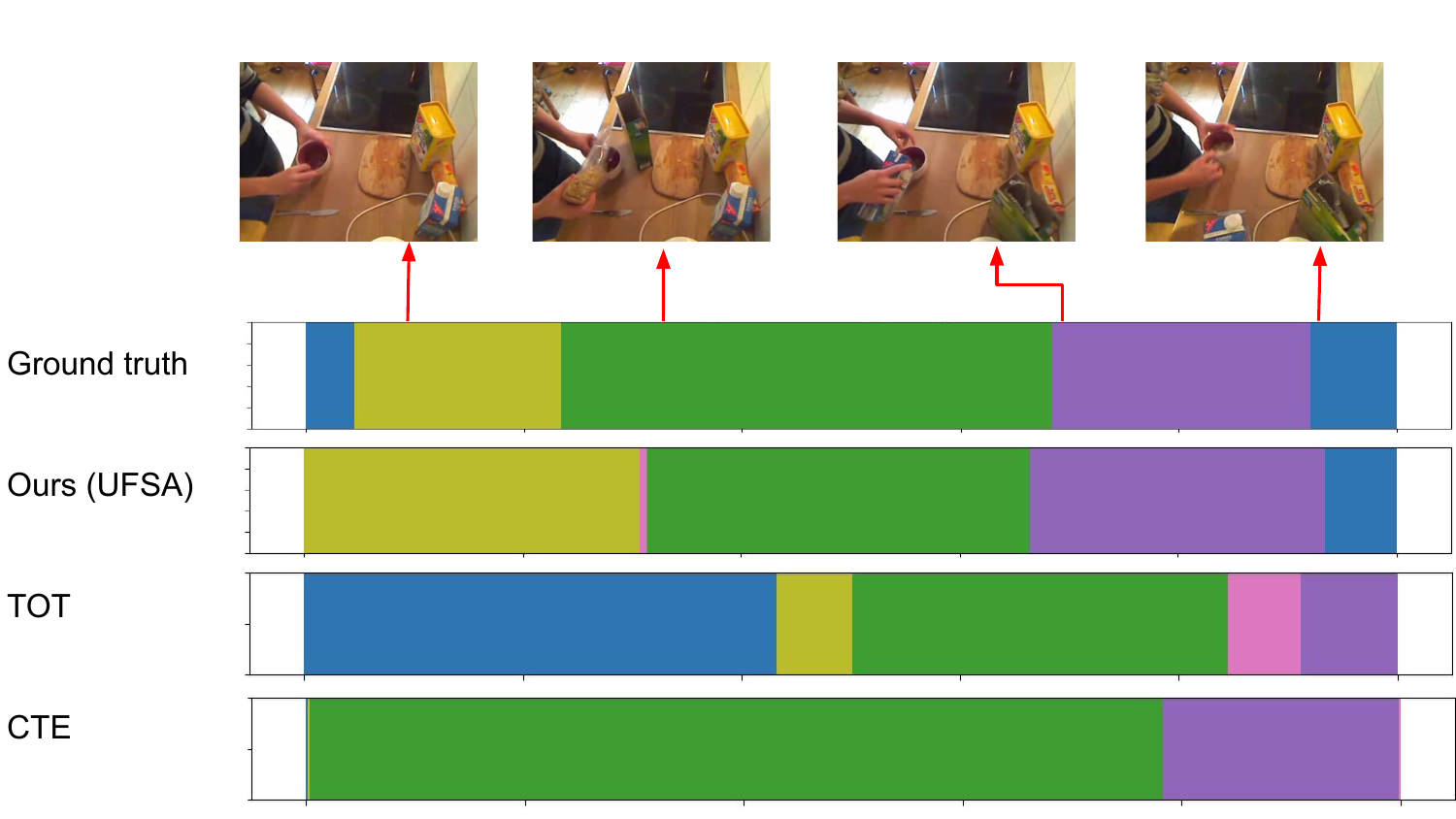}
         \caption{Breakfast (\emph{P13\_webcam01\_P13\_cereals}).}
         \label{fig:qual_bf}
     \end{subfigure}
    \hfill
     \centering
     \begin{subfigure}[b]{0.49\textwidth}
         \centering
         \includegraphics[width=\textwidth, trim = 5mm 0mm 0mm 0mm]{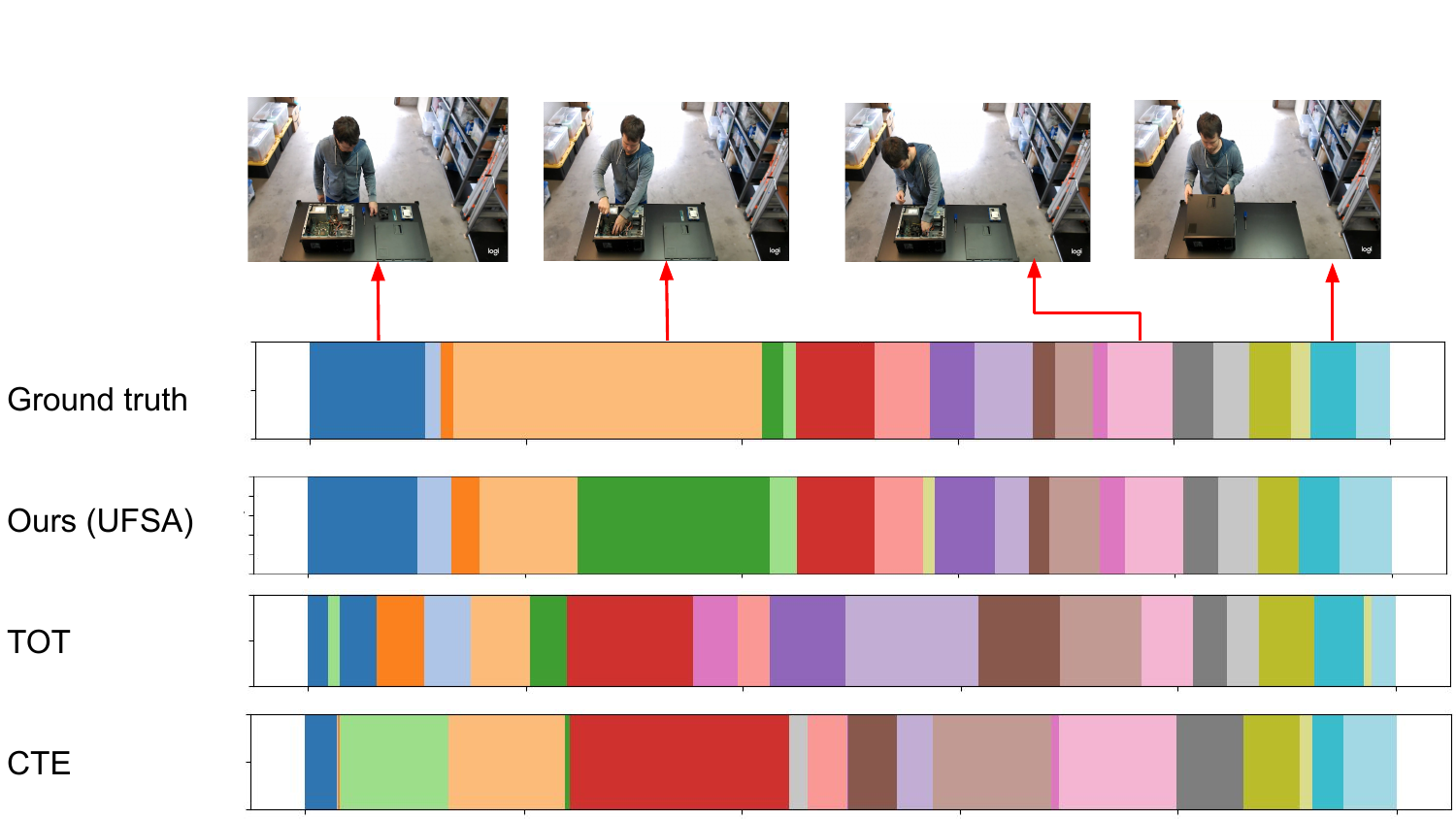}
         \caption{Desktop Assembly (\emph{2020-04-19\_17-24-35}).}
         \label{fig:qual_da}
     \end{subfigure}
     \caption{Segmentation results on (a) a YouTube Instructions video, (b) a Breakfast video, and (c) a Desktop Assembly video (\emph{`Orig'} set). }
     \label{fig:qual_segm}
\end{figure}

Moreover, we include in Fig.~\ref{fig:qual_segm} the segmentation results of our approach, TOT~\cite{kumar2022unsupervised}, and CTE~\cite{kukleva2019unsupervised} on other datasets, namely YouTube Instructions, Breakfast, and Desktop Assembly (\emph{Orig} set). It is evident from Fig.~\ref{fig:qual_segm} that our segmentation results are consistently closer to the ground truth than those of TOT~\cite{kumar2022unsupervised} and CTE~\cite{kukleva2019unsupervised}. 

Nevertheless, our approach has a limitation in handling repetitive actions. For example, let us look at the \emph{`Cut'} action in Fig.~\ref{fig:qual_perm}, which includes \emph{`Cut tomato'}, \emph{`Cut cucumber'}, \emph{`Cut cheese'}, and \emph{`Cut lettuce'} and hence occurs multiple times in the videos, our approach merges the multiple occurrences into a large segment (highlighted by blue boxes) since it assumes each action can happen only once. In addition, although TOT~\cite{kumar2022unsupervised} has the same drawback, our combined segments are closer to the ground truth.

\subsection{Ablation with MLP encoder}
\label{sec:supp_mlp}
\begin{table}[t]

    \centering
    \footnotesize
    \begin{tabular}{||c|c|c|c|c||}
    \hline
    & \textbf{Encoder} & \textbf{Decoder} &\textbf{MOF}  &\textbf{F1}\\
    \hline 
    \multirow{4}{*}{\rotatebox[origin=c]{90}{\textbf{\footnotesize{Eval}}}}
    & MLP & - & 47.4 & 31.8 \\
    & Transformer & - & 43.1 & 34.4  \\
    & MLP & Transformer & \underline{\textit{47.8}} & \underline{\textit{34.8}} \\
    &  \cellcolor{babyblueeyes} Transformer &  \cellcolor{babyblueeyes} Transformer &  \cellcolor{babyblueeyes} \textbf{55.8} &  \cellcolor{babyblueeyes} \textbf{50.3} \\
    \hline
    \multirow{4}{*}{\rotatebox[origin=c]{90}{\textbf{\footnotesize{YTI}}}}
    & MLP & - & 40.6 & 30.0 \\
    & Transformer & - & 42.8 & 30.2  \\
    & MLP & Transformer & \underline{\textit{43.2}} & \underline{\textit{30.5}} \\
    &  \cellcolor{babyblueeyes} Transformer &  \cellcolor{babyblueeyes} Transformer &  \cellcolor{babyblueeyes} \textbf{49.6} &  \cellcolor{babyblueeyes} \textbf{32.4} \\
    \hline
    \end{tabular}
    
    \caption{Ablation with MLP encoder. Best results are in \textbf{bold}, while second best ones are \underline{\textit{underlined}}.}
    
    \label{tab:supp_mlp}
\end{table}

We now perform an ablation study by using MLP encoder (instead of transformer encoder). Tab.~\ref{tab:supp_mlp} presents results on 50 Salads (\emph{Eval} granularity) and YTI datasets. From the results, transformer encoder alone performs similarly as MLP encoder alone (i.e., TOT). Next, MLP encoder+transformer decoder yields small improvements over TOT as features extracted by MLP encoder do not capture contextual cues that are useful for transformer decoder. Lastly, large improvements over TOT are achieved when transformer encoder is used jointly with transformer decoder (i.e., our complete model).

\subsection{Using $\boldsymbol{A}$ in segment-/alignment-level modules}
\label{sec:supp_a}
\begin{table}[t]

    \centering
    \footnotesize
    \begin{tabular}{||c|c|c|c||}
    \hline
    & \textbf{Method} &\textbf{MOF} &\textbf{F1}\\
    \hline 
    \multirow{3}{*}{\rotatebox[origin=c]{90}{\textbf{\footnotesize{Eval}}}}
    & Frame & 43.1 & 34.4 \\
    & Frame+Segment & \underline{\textit{43.3}} & \underline{\textit{37.8}} \\
    & Frame+Segment+Alignment & \textbf{46.1} & \textbf{45.2} \\
    \hline
    \multirow{3}{*}{\rotatebox[origin=c]{90}{\textbf{\footnotesize{YTI}}}}
    & Frame & 42.8 & \underline{\textit{30.2}} \\
    & Frame+Segment & \underline{\textit{43.3}} & \textbf{30.5} \\
    & Frame+Segment+Alignment & \textbf{44.3} & 29.4 \\
    \hline
    \end{tabular}
    
    \caption{Using $\boldsymbol{A}$ in segment-/alignment-level modules. Best results are in \textbf{bold}, while second best ones are \underline{\textit{underlined}}.}
    
    \label{tab:supp_a}
\end{table}

In this section, we repeat the ablation experiment in Tab.~1 of the main paper but we use the fixed-order prior $\boldsymbol{A}$ (instead of the permutation-aware prior $\boldsymbol{T}$) in segment-/alignment-level modules. Tab.~\ref{tab:supp_a} shows results on 50 Salads (\emph{Eval} granularity) and YTI datasets. It can be seen from the results that using $\boldsymbol{A}$ in segment-/alignment-level modules improves results of frame-level module only, however, the improvements are smaller than those of using $\boldsymbol{T}$ (see Tab.~1 of the main paper).

\subsection{Comparisons with ABD~\cite{du2022fast}}
\label{sec:abd}
\begin{table}[t]

    \centering
    \footnotesize
    \begin{tabular}{||c|c|c|c||}
    \hline
    & \textbf{Method} &\textbf{MOF}  &\textbf{F1}\\
    \hline
    \multirow{3}{*}{\rotatebox[origin=c]{90}{\textbf{\footnotesize{Eval}}}}
    & $^\star$ABD~\cite{du2022fast} & \textbf{71.4} & - \\
    & $^\dagger$ABD~\cite{du2022fast} & 34.2 & \underline{\textit{32.8}} \\
    & \cellcolor{babyblueeyes} $^\dagger$Ours (UFSA) & \cellcolor{babyblueeyes} \underline{\textit{55.8}} & \cellcolor{babyblueeyes} \textbf{50.3}  \\
    \hline
    \multirow{3}{*}{\rotatebox[origin=c]{90}{\textbf{\footnotesize{YTI}}}}
    & $^\star$ABD~\cite{du2022fast} &  \textbf{67.2} &  \textbf{49.2} \\
    & $^\dagger$ABD~\cite{du2022fast} & 29.4 &29.4\\
    & \cellcolor{babyblueeyes} $^\dagger$Ours (UFSA) & \cellcolor{babyblueeyes}  \underline{\textit{49.6}} & \cellcolor{babyblueeyes}  \underline{\textit{32.4}} \\
    \hline
    \multirow{3}{*}{\rotatebox[origin=c]{90}{\textbf{\footnotesize{Breakfast}}}}
    & $^\star$ABD~\cite{du2022fast} &  \textbf{64.0} &  \textbf{52.3}  \\
    & $^\dagger$ABD~\cite{du2022fast} & 23.6 & 21.7  \\
    & \cellcolor{babyblueeyes} $^\dagger$Ours (UFSA) & \cellcolor{babyblueeyes}  \underline{\textit{52.1}}  & \cellcolor{babyblueeyes}  \underline{\textit{38.0}} \\
    \hline
    \multirow{3}{*}{\rotatebox[origin=c]{90}{\textbf{\footnotesize{Orig}}}}
    & $^\star$ABD~\cite{du2022fast} &   \underline{\textit{63.3}}&   \underline{\textit{60.9}} \\
    & $^\dagger$ABD~\cite{du2022fast} & 15.5 & 11.0 \\
    & \cellcolor{babyblueeyes} $^\dagger$Ours (UFSA) & \cellcolor{babyblueeyes} \textbf{71.2}  & \cellcolor{babyblueeyes} \textbf{72.2} \\
    \hline
    \multirow{3}{*}{\rotatebox[origin=c]{90}{\textbf{\footnotesize{Extra}}}}
    & $^\star$ABD~\cite{du2022fast} &  \textbf{60.8} &  \textbf{57.1}  \\
    & $^\dagger$ABD~\cite{du2022fast} & 12.0 & 10.6  \\
    & \cellcolor{babyblueeyes} $^\dagger$Ours (UFSA) &\cellcolor{babyblueeyes}  \underline{\textit{58.6}}  & \cellcolor{babyblueeyes}   \underline{\textit{55.9}} \\
    \hline
    \end{tabular}
    
    \caption{Comparisons with ABD~\cite{du2022fast}. Note that $^\star$ denotes video-level results, whereas $^\dagger$ denotes activity-level results. Best results are in \textbf{bold}, while second best ones are \underline{\textit{underlined}}.}
    \label{tab:abd}
\end{table}
Our method addresses the problem of activity-level segmentation, which jointly segments and clusters frames across all input videos. A related problem is video-level segmentation, which aims to segment a single input video only. Video-level segmentation is a sub-problem of activity-level segmentation and in general easier than activity-level segmentation. In this section, we evaluate the performance of a recent video-level segmentation method, i.e., ABD~\cite{du2022fast}, for the task of activity-level segmentation. Firstly, for each input video, we run ABD~\cite{du2022fast} to obtain its video-level segmentation result. We then represent each segment in the result by its prototype vector, which is the average of feature vectors of frames belonging to that segment. Next, we perform K-Means clustering (K is set as the ground truth number of actions available in the activity) on the entire set of prototype vectors from all input videos to obtain the activity-level segmentation result, which we evaluate in Tab.~\ref{tab:abd}. From the results, it can be seen that $^\dagger$UFSA outperforms $^\dagger$ABD~\cite{du2022fast} in the activity-level setting on all metrics and datasets. A more advanced clustering method which incorporates temporal information can be used instead of K-Means, however, it is out of the scope of our work. In addition, the video-level results of $^\star$ABD~\cite{du2022fast} are mostly better than the activity-level results of $^\dagger$UFSA (except for Desktop Assembly - \emph{Orig}), which is due to fine-grained video-level Hungarian matching~\cite{vidalmata2021joint}.

\subsection{Implementation Details}
\label{sec:implementation}

\noindent\textbf{Hyperparameter Settings.}
Tab.~\ref{tab:hyperparameters} presents a summary of our hyperparameter settings. For the temporal optimal transport problem in our frame-level prediction module and frame-to-segment alignment module, we follow the same hyperparameter settings used in TOT~\cite{kumar2022unsupervised}, including $\rho$ and number of Sinkhorn-Knopp iterations. We keep the feature dimension $d$ the same as TOT~\cite{kumar2022unsupervised}. We use a single video, including all frames, per batch. In addition, for our transformer encoder and transformer decoder, we follow the same hyperparameter settings used in UVAST~\cite{behrmann2022unified}, including encoder dropout ratio and decoder dropout ratio. We set the temperature $\tau = 0.1$ (same as TOT~\cite{kumar2022unsupervised}) in Sec.~3.1 of the main paper and the temperature $\tau^\prime = 10^{-3}$ (same as UVAST~\cite{behrmann2022unified}) in Sec.~3.3 of the main paper.

\noindent\textbf{Computing Resources.} All of our experiments are conducted with a single Nvidia A100 SXM4 GPU on Lambda Cloud.

\begin{table*}[ht!]
\begin{minipage}[t]{1.0\linewidth}
\centering
\footnotesize
{
\begin{tabular}{l|l}

\specialrule{1pt}{1pt}{1pt}

\textbf{Hyperparameter} &  \textbf{Value}\\

\midrule

Temperature ($\tau$) & $0.1$ \\

Rho ($\rho$) &  $0.07$ (\textbf{E}), $0.08$ (\textbf{M}), $0.08$ (\textbf{Y}), $0.05$ (\textbf{B}), $0.07$ (\textbf{O}), $0.07$ (\textbf{A})    \\
Number of Sinkhorn-Knopp iterations & $3$\\

Feature dimension ($d$) & $30$ (\textbf{E}), $30$ (\textbf{M}), $200$ (\textbf{Y}), $40$ (\textbf{B}), $30$ (\textbf{O}), $30$ (\textbf{A}) \\

Batch size & $1$\\

Learning rate  & $10^{-3}$\\
Weight decay & $10^{-5}$\\

Number of encoder layers & $2$\\
Number of decoder layers & $2$\\

Encoder dropout ratio & $0.3$\\
Decoder dropout ratio & $0.1$\\

Temperature ($\tau^\prime$) & $10^{-3}$ \\
\specialrule{1pt}{1pt}{1pt}
\end{tabular}
}
\caption{Hyperparameter settings. \textbf{E} denotes 50 Salads (\emph{Eval} granularity), \textbf{M} denotes 50 Salads (\emph{Mid} granularity), \textbf{Y} denotes YouTube Instructions, \textbf{B} denotes Breakfast, \textbf{O} denotes Desktop Assembly (\emph{Orig} set), and \textbf{A} denotes Desktop Assembly (\emph{Extra} set).}
\label{tab:hyperparameters}

\end{minipage}

\end{table*}

\subsection{Societal Impacts}
\label{sec:societal}

Our approach facilitates video recognition model learning without action labels, with potential applications in frontline worker training and assistance. Models generated from expert demonstration videos in various domains could offer guidance to new workers, improving the standard of care in fields such as medical surgery. However, at the same time, video understanding algorithms in surveillance applications may compromise privacy, even if it enhances security and productivity. Thus, we urge caution in the implementation of such technologies and advocate for the development of appropriate ethical guidelines.

{\small
\bibliographystyle{ieee_fullname}
\bibliography{references}
}

{\small
\bibliographystyle{ieee_fullname}
\bibliography{references}
}

\end{document}